\documentclass[11pt]{article}
\usepackage[letterpaper,margin=1in]{geometry}
\usepackage[T1]{fontenc}
\usepackage[utf8]{inputenc}
\usepackage{lmodern}
\usepackage{amsmath,amsfonts,booktabs,graphicx,multirow,placeins}
\usepackage{xurl,tikz,pifont}
\usetikzlibrary{positioning,arrows.meta}
\usepackage[authoryear,round]{natbib}
\usepackage{microtype}
\usepackage{hyperref}
\hypersetup{hidelinks,pdftitle={TPBench: A Turning-Point Benchmark for Dialogue Compression},pdfauthor={Minji Park, Seunghyun Yoon, Hyuk Lim}}
\newcommand{\yes}{\ding{51}}
\newcommand{\no}{\textcolor{red!70!black}{\ding{55}}}
\title{TPBench: A Turning-Point Benchmark for Dialogue Compression}
\author{
Minji Park, Seunghyun Yoon, Hyuk Lim  \\[0.4em]
Korea Institute of Energy Technology (KENTECH) \\
\texttt{\{minjipark, syoon, hlim\}@kentech.ac.kr}
}
\date{}
\begin{document}
\maketitle

\begin{abstract}
A compressor can keep the facts of a dialogue and still drop the turn that
changed them. A user corrects a price, reverses a choice, or adds a
constraint. We call this failure turning-point eviction. One overall
retention score hides it, because that score mixes what the user first
wanted with what the user wants now.

We introduce TPBench, which evaluates three complementary information targets
at shared nominal retention budgets. P1 asks for the user's initial goal. P2 asks for the current value
of a slot the user revised. P3 asks for both, in dialogues with a late annotated
slot update. The current-value answers come from the human dialogue-state annotations of
MultiWOZ and SGD. The initial-goal answer is the first sentence of the first
user turn. Neither requires new crowdsourcing.

The probe-specific evaluations rank compression methods differently. On the
joint probe at a retained fraction of $0.30$, every tested compressed method
remains below full context with the main Llama reader. Deleting the turn
that carries the update sharply lowers current-value accuracy, while
deleting one matched irrelevant turn leaves it unchanged. A Mistral reader repeats the P2/P3 rankings and the joint-probe gap.
Current-value recovery is tested on an additional corpus, LongMemEval-KU, and
on Chinese RiSAWOZ: full context has the highest accuracy, and recency has the
highest compressed-method mean in both evaluations.
\end{abstract}

\section{Introduction}
\label{sec:intro}

A user asks an assistant to find a hotel in the east. Later, the user changes
the price to under \$50. A compressor may preserve the hotel request and drop
the price change. Although the compressed dialogue remains topically relevant, it no longer contains the evidence needed to determine the current price constraint. We refer to this failure as \emph{turning-point
eviction}.

The initial request and the updated price occupy different parts of the
dialogue. The initial request describes
the task. The updated price records the latest value of a \emph{slot}, or named
field. Price, departure time, and destination are examples of slots.

Within each comparison, the same language model reads each compressed dialogue. We call it
the \emph{reader}. The reader answers a question using only the context that a
compression method retained. Its answer reveals which information remains
usable after compression.

TPBench contains three probes:

\begin{itemize}\itemsep2pt
    \item \textbf{P1 (initial goal)} asks what the user originally wanted.
    \item \textbf{P2 (current value)} asks for the final value of a slot that
    changed during the dialogue.
    \item \textbf{P3 (initial goal and late update)} asks for both answers in
    dialogues where the final annotated slot update occurs in the second half.
\end{itemize}

Figure~\ref{fig:worked_example} shows why the three results are reported
separately. A context that keeps only the first turn can answer P1. A context
that keeps only the later price statements can answer P2. P3 requires both
parts of the dialogue.

\begin{figure}[!htbp]
\centering
\centering
\begin{tikzpicture}[font=\footnotesize,
  turn/.style={draw=black!55,rounded corners=2pt,inner sep=3.5pt,text width=53mm,align=left},
  point/.style={turn,draw=black,very thick,fill=black!8},
  ctx/.style={draw=black!55,rounded corners=2pt,inner sep=3.5pt,text width=53mm,align=left},
  tag/.style={font=\scriptsize\itshape,black!60},
  hd/.style={font=\small\bfseries}]
\node[turn] (t0) at (0,0)
  {\textbf{T0} \textsc{user}: Find a hotel in the east.\\[1pt]\tikz\node[tag]{initial goal};};
\node[turn,below=1.8mm of t0.south west,anchor=north west] (t3)
  {\textbf{T3} \textsc{user}: A moderate price is fine.\\[1pt]\tikz\node[tag]{earlier value};};
\node[point,below=1.8mm of t3.south west,anchor=north west] (t6)
  {\textbf{T6} \textsc{user}: Actually, make it under \$50.\\[1pt]\tikz\node[tag,text=black]{\textbf{turning point}};};
\node[turn,below=1.8mm of t6.south west,anchor=north west] (t7)
  {\textbf{T7} \textsc{system}: I found one under \$50.\\[1pt]\tikz\node[tag]{answer repeated};};
\node[hd,above=2mm of t0.north] {A dialogue with one revision};
\node[ctx,right=6mm of t0.north east,anchor=north west] (c1)
  {\textbf{keeps T0, drops T6 and T7}\\[2pt]\yes\ P1 initial goal\hfill\no\ P2 current value};
\node[ctx,below=1.8mm of c1.south west,anchor=north west] (c2)
  {\textbf{drops T0, keeps T6 and T7}\\[2pt]\no\ P1 initial goal\hfill\yes\ P2 current value};
\node[ctx,below=1.8mm of c2.south west,anchor=north west,draw=black,very thick] (c3)
  {\textbf{keeps T0 and T6}\\[2pt]\yes\ P3 asks for both at once};
\node[hd,above=2mm of c1.north] {What a compressed context supports};
\end{tikzpicture}
\caption{An illustrative dialogue separates the three probes. P1 needs the
initial goal, P2 needs the current value, and P3 needs both. On the left, the
heavy box is the turning point, the user turn that revises the slot. On the
right, each box is one compressed context; a check mark means the probe can be
answered from it and a cross means it cannot. The heavy box on the right is the
only context that answers P3.}
\label{fig:worked_example}
\end{figure}

A \emph{turn} is one user or system message, indexed from zero. We use
\emph{turning point} operationally for a user turn with an annotated slot
update. An annotation change need not be a semantic revision: P2 removes
changes identified by its alias heuristic, while P3 retains late annotation
changes and reports an alias-exclusion sensitivity analysis.

Schema-Guided Dialogue (SGD)~\citep{rastogi2020sgd} and
MultiWOZ 2.2~\citep{budzianowski2018multiwoz,zang2020multiwoz22}
make these questions reproducible. Both corpora contain multi-turn dialogues
from several service domains. Their annotations record active slot values as
the conversation changes. TPBench derives questions from these records. It
then evaluates every compression method with the same reader and scoring
rules.

The benchmark serves two purposes. First, it identifies which part of a
dialogue a method tends to preserve. Second, it measures whether both parts
remain available when an application needs the original task and the latest
constraint together. TPBench supplies probe-specific comparisons of released compression
pipelines for these complementary requirements.

\paragraph{Contributions}
\begin{itemize}\itemsep1pt
\item We separate dialogue-compression evaluation into three targets:
recovering the user's initial goal, recovering the current value after a
revision, and recovering both after a late update.
\item We build TPBench from the SGD and MultiWOZ dialogue-state annotations.
Probe construction and scoring are deterministic, and no new crowdsourced
labels are needed.
\item Across seven turn selectors and four KV-cache methods, the preferred
method differs across probe-specific evaluations. In the main Llama
evaluation, joint recovery remains substantially below full context for every
tested turn selector.
\item We validate the current-value probe with matched deletion controls, and
we test P2/P3 with a second reader and current-value recovery with an
additional corpus and Chinese dialogue.
\end{itemize}

\section{Related work}
\label{sec:related}

\paragraph{Long-context compression methods}
Prompt-side compression shortens the text sent to a reader. LLMLingua-2
scores token importance~\citep{pan2024llmlingua2}, and LongLLMLingua extends
the idea to long inputs by reordering and budgeting across
segments~\citep{jiang2024longllmlingua}. ICAE represents context in compact
memory slots~\citep{ge2024icae}. MMR selects relevant and diverse
content~\citep{carbonell1998mmr}, an objective that predates neural readers
and is still used to rank passages alongside lexical scoring
functions~\citep{robertson2009bm25} and dense sentence
encoders~\citep{devlin2019bert,reimers2019sbert,chen2024bgem3}.
\citet{li2024promptcompsurvey} survey the area, and \citet{nagle2024promptlimits}
give a rate-distortion account of what any black-box prompt compressor can
achieve. These methods differ in the signal they score: token probability,
embedding similarity, or attention weight. Each signal answers a different
question about a turn. Under a diversity criterion, the turn that states the user's opening request
can lose weight once later turns repeat the topic. Under a token-level
importance criterion, a brief revision can receive little weight. Our measurements make this concrete: the selector that
recovers the initial goal best is not the selector that recovers the current
value best, on either dataset.

KV-cache compression reduces the internal state that a
transformer~\citep{vaswani2017attention} retains during decoding.
Attention scores guide H2O~\citep{zhang2023h2o} and
ScissorHands~\citep{liu2023scissorhands}. Other retention schemes include
SnapKV~\citep{li2024snapkv}, PyramidKV~\citep{cai2025pyramidkvdynamickvcache},
and StreamingLLM~\citep{xiao2024streamingllm}, whose attention-sink argument
rests on the position of a token in the sequence rather than its content. KIVI
quantizes the cache~\citep{liu2024kivi}. ChunkKV prunes groups of cache
entries~\citep{liu2025chunkkv}. A more recent line allocates the budget adaptively rather than uniformly.
Ada-KV varies the budget across attention heads~\citep{feng2025adakv} and
DynamicKV varies it across layers~\citep{zhou2025dynamickv}. KeyDiff evicts by
key similarity~\citep{park2025keydiff}, and FastKV separates context reduction
from cache compression~\citep{jo2026fastkv}. This family and the prompt-side
family retain different objects, a fraction of turns against a fraction of
cache entries, so a shared retained fraction does not make their budgets
equivalent. TPBench evaluates representatives from both families under the
same probe questions and reader model, supporting descriptive accuracy
comparisons across the released pipelines despite their different budget units.

\paragraph{Where information sits in a long input}
Position is known to govern what a reader recovers. \citet{liu2024lostmiddle}
show that accuracy is highest when the relevant passage lies at either end of
the input and lowest in the middle, and \citet{hsieh2024foundmiddle} trace part of
that pattern to positional attention bias. This literature treats position as a
nuisance to be corrected. A dialogue turns it into a substantive problem,
because the initial goal is anchored near the beginning while a revised value
may be established later. In P3 the final update is constrained to the second
half of the dialogue. A compressor that keeps one end therefore does not fail at
random: it fails on a particular probe, which is what TPBench measures.
Several lines of work treat a point of change as the unit that structures a
discourse: event boundaries~\citep{zacks2007}, dialogue-act
modeling~\citep{stolcke2000dialogue}, and topic segmentation in
text~\citep{hearst1997texttiling} and in
dialogue~\citep{park2023dialtopicseg}. A turn-level importance score need not explicitly represent the role of an
update in changing the dialogue state.

\paragraph{Long-context and dialogue evaluation benchmarks}
LongBench~\citep{bai2024longbench} and $\infty$Bench~\citep{zhang2024infty}
test a broad range of long-context abilities. LongGenBench targets long-form
generation~\citep{liu2024longgenbench}. CLongEval covers
Chinese~\citep{qiu2024clongeval}.
MT-Bench evaluates multi-turn instruction
following~\citep{zheng2023mtbench}. LongMemEval tests long-term memory across
sessions~\citep{wu2025longmemeval}, and long-term dialogue datasets supply the
conversational setting for such memory~\citep{xu2022msc,smith2020bst,xu2022longtimenosee,jang2024mixedsession}. These benchmarks provide broad task
coverage, but do not explicitly separate initial-goal recovery from later
current-value recovery at shared nominal retention budgets. TPBench
measures these targets separately and adds a probe that requires both at once.

\paragraph{Dialogue-state and conversational benchmarks}
SGD~\citep{rastogi2020sgd} and MultiWOZ
2.2~\citep{budzianowski2018multiwoz,zang2020multiwoz22} are widely used for
dialogue-state tracking. The task was organized for years around the Dialog
State Tracking Challenges~\citep{henderson2014dstc2,williams2016dstc}. It has
been addressed by generation-based trackers such as
TRADE~\citep{wu2019trade} and, more recently, by prompted language
models~\citep{hudecek2023llmtod}. Standard DST
evaluates the full state against a schema, and its scoring conventions have
themselves been questioned for hiding turn-level
behavior~\citep{dey2022fairdst}. TPBench uses changes in that state to
create questions. The reader sees the dialogue and a natural-language
question. It does not receive the schema. This difference matters for what a
failure means: a DST model that misses an update has misread the dialogue,
whereas a reader in our setting may never have been given the turn that
carries the update. The turn-selection comparisons therefore hold the reader fixed while
varying the retained dialogue text. Cache-compression comparisons instead
vary the reader's retained attention state.
Information-state accounts of dialogue treat the state as revised turn by
turn~\citep{larsson2000isu}. Grounding accounts describe how participants
register such revisions~\citep{clark1991grounding}. Both motivate treating a revision as a state-changing event whose evidential
role differs from that of the surrounding turns.
Multi-Session Chat~\citep{xu2022msc} and Blended Skill
Talk~\citep{smith2020bst} cover less structured conversations. Their records do
not provide the slot-level updates required by the present construction.

\section{TPBench}
\label{sec:tpbench}
\label{sec:probes}

Each TPBench example contains a dialogue, a question, and a reference answer.
A turn selector shortens the dialogue text; a KV-cache method reduces the
reader's retained attention state. The reader answers using the resulting
context. Within each comparison, methods share the reference answers and
reader model.

The two targets differ in what a compressor has to keep, not only in how hard
they are. Dialogue-theoretic accounts treat the information state as revised
turn by turn \cite{larsson2000isu}, a view also adopted in dialogue state tracking. We distinguish information anchored in the initial user turn from information determined by the latest slot update. P1 evaluates recovery of the former, P2 evaluates the latter, and P3 requires both.

P1 fixes its reference to the first user turn, whereas P2 follows the final
annotated update to the queried slot. A method that keeps the anchor turn can
score well on P1 while discarding the evidence needed by P2. The paired
deletions in Section~\ref{sec:finding3} measure the reader's dependence on the
update turn and on the complete set of answer-bearing turns.
Table~\ref{tab:probe_overview} states the question and the reference answer for
each probe.

\begin{table}[!htbp]
\centering\footnotesize\setlength{\tabcolsep}{3.5pt}
\caption{The three TPBench probes. The Question column paraphrases each prompt.
Appendix~A gives the prompts verbatim.}
\label{tab:probe_overview}
\begin{tabular}{@{}p{0.08\linewidth}p{0.39\linewidth}p{0.45\linewidth}@{}}
\toprule
Probe & Question & Reference answer \\
\midrule
P1 & What was the user's initial goal? & first sentence of the first user turn \\
P2 & What is the current value of slot $s$? & value after the final update to $s$ \\
P3 & What is the task, and what is current? & first-turn goal and value after the final update \\
\bottomrule
\end{tabular}
\end{table}

\paragraph{P1: initial goal}
The reader answers, ``What was the user's initial goal at the beginning of this conversation?''
The reference is the first sentence of the first user turn. We lowercase
alphanumeric tokens, discard tokens shorter than two characters, and remove a
predefined stopword list. The remaining distinct tokens are the content words. The strict score requires at least two shared content words.
The loose score requires at least one. Token $F_1$ gives a graded view of the
same overlap.

\paragraph{P2: current value}
The reader is given a slot name, such as \texttt{train-departure}, and asked for
its current value. The reference comes from the last annotated update to that
slot. Normalization handles common equivalents such as \texttt{12 pm} and
\texttt{12:00}. The strict score requires equality after normalization. The
loose score also accepts a longer prediction that contains the complete
reference at phrase boundaries.

\paragraph{P3: initial goal and late update}
P3 scores recovery of the initial-goal reference and current-value reference
in one answer. It uses dialogues
with at least ten turns. The final update must occur at index six or later. It
must also occur in the second half of the dialogue. Both P3 scores require one
shared content word for the initial goal. P3-strict requires the complete
current value at phrase boundaries. P3-loose accepts that match or at least
$\max(1,\lfloor |V|/2\rfloor)$ shared value content words, where $V$ is the
set of reference value content tokens. The saved prompt requests ``task; slot: value'' inside a JSON \texttt{value}
field. It asks for ``the user's task in this conversation,'' whereas the goal
reference remains the first sentence of the first user turn. Unlike P1, this
wording does not explicitly request the initial goal; in multi-task dialogues
it can invite a current or overall task description. P3 therefore scores the
goal segment more loosely than P1 does.
The P3 scorer parses that field into a goal segment and a \texttt{slot: value}
segment. It applies the goal criterion to the goal segment and the value
criterion to the value segment. An answer the parser cannot read scores zero.
Section~\ref{sec:p3_segment_sensitivity} reports the alternative of scoring the
whole field at once, together with the share of answers that parse.

P1 and P2 use different questions and different scoring rules. Their absolute
accuracy values should therefore be read within each probe. The ranking of
methods within P1 can be compared with the ranking within P2. P3 directly
measures joint recovery on one set of examples. Strict and loose P1 scores
report multiword and single-word overlap, respectively. P3-strict uses its own
component thresholds; it is not the conjunction of standalone P1-strict and
P2-strict. The probes use separately sampled sets with different eligibility
rules. Methods are paired within each probe; P1/P2 ranking comparisons also
reflect differences between their sampled sets.

\section{Constructing the examples}
\label{sec:gold}

SGD and MultiWOZ 2.2 provide dialogue-state tracking (DST) annotations that
record active slot values throughout a conversation. These annotations are
produced and corrected by people, and MultiWOZ 2.2 in particular revises the
state labels of earlier releases. Their multi-domain dialogues contain many
revisions, and the state records make those revisions identifiable without
collecting new labels. The two corpora also
use different collection procedures and annotation formats. Using both datasets tests whether the observed effects persist across different annotation formats and data-collection procedures.

\paragraph{Initial-goal examples}
P1 takes the first sentence of the first user turn. This reference is fixed
by position rather than by an annotation, and it selects one initial-goal
reference for each sampled dialogue. We use the same sentence
when later slot values change.

\paragraph{Current-value examples}
A source question is a stored slot-update question derived from a dialogue;
several source questions can refer to one source dialogue. P2 construction
follows five steps.
\begin{enumerate}\itemsep1pt
    \item Compare adjacent dialogue states at each user turn.
    \item Record every slot whose annotated value changes.
    \item Normalize the old and new values and discard pairs accepted by the
    corpus-specific alias-equivalence heuristic.
    \item Require the normalized new value to occur as a substring of the dialogue text.
    \item Keep only the final update to the selected slot.
\end{enumerate}
The fourth step guards against values that exist only in the annotation. The
fifth step prevents an earlier value from being labeled current. Each seed draws $200$ source questions and then applies these five rules to
them, so a cell keeps whatever survives. Effective sample sizes are
$\{128,125,132\}$ for SGD and $\{161,168,168\}$ for MultiWOZ at seeds
$\{42,43,44\}$, respectively. Over the whole source pool the final-update rule
leaves $791$ SGD and $664$ MultiWOZ candidates, and the alias rule reduces
these to $623$ and $628$. The value-occurrence rule removes no further item.

\paragraph{Late updates for P3}
P3 selects final annotated slot updates and applies its length and position
rules. Unlike P2, its builder does not apply the alias-equivalence or
value-occurrence filters. Its eligible pools contain $452$ SGD and $277$
MultiWOZ examples. Each seed draws $200$ examples from these eligible pools,
so every primary P3 cell has exactly $200$ examples. P3 thus measures joint
recovery after late annotation changes, including some changes that the P2
alias heuristic treats as equivalent. Section~\ref{sec:p3_alias_sensitivity}
reports a sensitivity analysis excluding those pairs from the saved outputs.
These distinct sampling and eligibility rules explain the P2/P3 sample-size
difference; P3 is not a late-update subset of a sampled P2 set.

\paragraph{Update turns and answer-bearing turns}
The \emph{update turn} is the user turn where the annotation changes to the
final value. An \emph{answer-bearing turn} is any user or system turn that
states the complete normalized reference. These sets can differ. In
Figure~\ref{fig:worked_example}, T6 is the update turn. Both T6 and T7 carry the
answer. The same phrase-boundary matcher locates answer-bearing turns and
scores loose P2 predictions.

\paragraph{Audit of the update-turn mapping}
The mapping follows the recorded conversion order of the source dialogues.
Every accepted update maps to a corresponding user turn. Repeated annotations
create more than one matching turn for $16$ MultiWOZ questions. The recorded
conversion order resolves those ties.

\paragraph{Matched deletion controls}
The first P2 deletion removes only the update turn. Its control removes one
irrelevant turn from the same speaker. Word length and dialogue position choose
the closest match. The second deletion removes the update turn and every
answer-bearing turn. Its control matches both the number of turns and the
speaker composition. Matching succeeds for all $128$ SGD examples and for
$159$ of the $161$ MultiWOZ examples.

\paragraph{Leak checks}
We check two potential sources of evaluation leakage. First, the P1 reference is the first sentence of turn~0, so any
selector that keeps the earliest turns retains the P1 reference by position.
\emph{Earliest turns} keeps turn~$0$ by rule and scores $0.480$ on SGD
P1-strict against $0.487$ for full context. P1 therefore measures whether a
selector keeps the opening of the dialogue. Second, the seed sample could over-represent dialogues with explicit revision
wording, which would favour methods keyed to correction cues. The audit compares the examples a seed actually evaluates against the rest of
the final-update pool of $791$ and $664$ candidates. On seed $42$, a fixed user-side
correction-cue regular expression fires on $28.1\%$ of the $128$ evaluated SGD
examples versus $21.3\%$ of the $663$ other final-update candidates
($+6.9$~pp), and on $32.3\%$ of the $161$ evaluated MultiWOZ examples versus
$28.8\%$ of the remaining $503$ ($+3.5$~pp). These complements include candidates removed by P2 filtering,
so the comparison combines sampling and eligibility effects.
\section{Experimental setup}
\label{sec:setup}

\paragraph{Turn selectors}
We evaluate seven methods that retain whole dialogue turns. Four selectors use
position alone. Recency keeps the latest turns. The earliest-turn baseline
keeps the first turns. Random selection samples turns uniformly. Uniform stride
spreads retained turns across the dialogue.

Three selectors use dialogue content. The main evaluation supplies fixed
contexts labeled H2O-inspired attention proxy, embedding maximal marginal
relevance (MMR), and LLMLingua-2. The released attention
proxy averages attention over layers and heads, sums it over query positions,
and averages within each turn~\citep{zhang2023h2o}. The released MMR
implementation uses a trade-off of $0.5$ and the mean turn embedding as its
query~\citep{carbonell1998mmr}. The free-form experiment of
Section~\ref{sec:fmts}, which we call FMTS, records GPT-2 as the feature model
for these two selectors. The feature model and the generation environment for
the main H2O and MMR caches were not recorded, so that configuration does not
establish what produced them.

The LLMLingua-2 turn selector uses binary word-retention labels from
\nolinkurl{llmlingua-2-xlm-roberta-large-meetingbank}, run at its own token rate
of $0.5$~\citep{pan2024llmlingua2}. That rate governs the label model and is
separate from the turn budget of the experiments. It averages those labels within each turn and keeps the highest-scoring turns.

The release includes fixed compressed contexts and saved reader outputs.
These support reconstruction of the reader inputs and deterministic rescoring;
they do not by themselves establish identical outputs from a new inference run.

The saved main full-context and position-only inputs retain
\texttt{[Tn] USER/SYSTEM} turn and speaker markers. The H2O/MMR/LLMLingua-2
inputs concatenate selected turn texts without those markers. Together with
the floor/round budget difference below, this means the comparisons describe
released pipelines at shared nominal budgets; they do not isolate selection
logic from rendering and rounding choices.

\paragraph{KV-cache methods}
KV-cache compression leaves the displayed dialogue intact. It changes the
reader's stored attention keys and values. ChunkKV groups cache entries into
chunks before pruning~\citep{liu2025chunkkv}. It is the one cache method we
evaluate on P3, in Table~\ref{tab:p3_composite}. SnapKV, PyramidKV, and StreamingLLM are evaluated on P2.

The budget $r$ denotes the target fraction retained by a method. A main-pool
positional selector keeps
$k=\max(1,\operatorname{round}(n_{\mathrm{turns}}r))$ dialogue turns. The
LLMLingua-2 instead keeps
$k=\max(1,\lfloor n_{\mathrm{turns}}r\rfloor)$ turns, as do H2O and MMR. Thus the same nominal $r$ can produce a one-turn
budget difference. A KV method keeps the fraction
$r$ of the KV cache. 
The two method
families therefore share a retained-fraction convention. Their physical units
differ: retaining a fraction of turns need not retain the same fraction of tokens.

\paragraph{First-user-plus-recent baseline}
This selector retains the first USER turn and the latest $k-1$ other turns,
where $k=\max(1,\operatorname{round}(n_{\mathrm{turns}}r))$, in chronological
order. It uses only turn position and speaker identity. Its matched comparison
includes full context, earliest turns, recency, and uniform stride
(Section~\ref{sec:first_recent}).

\paragraph{Reader and samples}
The main reader is Llama-3.1-8B-Instruct~\citep{dubey2024llama3}. Decoding is
greedy. Turn-selection results use seeds $\{42,43,44\}$. P1 is evaluated at $r{=}0.30$. P2 uses $r\in\{0.10,0.30\}$ in the main tables
and $r\in\{0.50,0.70\}$ in the wider-budget comparison of
Section~\ref{sec:wider_ratios}. P3 uses $r{=}0.30$. Mistral-7B-Instruct
replicates P2 and P3, and the free-form experiment of Section~\ref{sec:fmts}
uses GPT-5.4-mini. Each seed begins with
$200$ sampled probe examples per dataset and probe. P2 filtering produces the
smaller effective counts reported above. Turn selectors within each cell share
the examples, questions, reader model, and probe-level decoding settings.
Their turn rounding and context rendering can differ as described above. The KV-cache evaluations use
the same reader model and probe questions, with cache-specific execution
settings. Full context uses the reader
without compression.

\paragraph{Scoring}
The headline P1--P3 scores use deterministic functions to compare the reader's
answers with the references. Table~\ref{tab:experiment_scope} lists the
reader, the sample, and the scoring rule of every comparison.

\begin{table}[!htbp]
\centering\footnotesize\setlength{\tabcolsep}{3.5pt}
\caption{Experiment scope. Main seeds sample probe sets. LongMemEval-KU and
RiSAWOZ instead keep the eligible set fixed and vary random selection. Turn
fractions and cache fractions have different physical units.}
\label{tab:experiment_scope}
\begin{tabular}{@{}p{0.19\linewidth}p{0.15\linewidth}p{0.25\linewidth}p{0.31\linewidth}@{}}
\toprule
Comparison & Reader & Examples and budget & Scoring \\
\midrule
Main turn selectors & Llama & Three sampled sets; P1/P3 $r=.30$, P2 $r=.10,.30$ & P1 lexical; P2 value; P3 segment-local joint \\
P1 semantic check & Llama answers, GPT-5.4-mini judge & Same three sampled sets; $r=.30$; six selectors & AI-assisted full equivalence \\
First user + recent & Llama & Same three sampled sets; $r=.30$ & Matched generation run; same probe scorers \\
Deletion controls & Llama & Seed-42 P2; paired turn deletion & P2 strict/loose; superseded-value diagnostic \\
Cache methods & Llama & Seed 42; cache $r=.30$ & Four methods on P2; ChunkKV on P3 \\
Cross-reader & Mistral & Same three P2/P3 sampled sets & Same probe scorers \\
LongMemEval-KU & Llama & All $72$ items; turn $r=.30$ & Corpus-specific current-value scorer \\
RiSAWOZ & Llama & All $207$ dialogues; turn $r=.30$ & Normalized exact current-value match \\
FMTS free-form & GPT-5.4-mini & $100/200/200$ items; turn $r=.10,.30$ & Bidirectional substring containment \\
\bottomrule
\end{tabular}
\end{table}

\FloatBarrier
\section{Findings}
\label{sec:findings}

\subsection{Rankings across the three probes}
\label{sec:f1}

The best compression method changes with the question. This section reports
the main scores, then asks whether a method falls behind because it throws the
answer away or because the reader misses an answer it kept.

\begin{table}[!htbp]
\centering\footnotesize\setlength{\tabcolsep}{3.5pt}
\caption[Main results at r=0.30.]{Main Llama results at $r{=}0.30$, averaged over
seeds $\{42,43,44\}$. P1/P3 use $200$ examples per dataset and seed; P2
uses $\{128,125,132\}$ SGD and $\{161,168,168\}$ MultiWOZ examples. P1 columns
show strict~/~loose scores, P2 is strict, and P3$_\mathrm{l}$ and
P3$_\mathrm{s}$ are the loose and strict joint scores of
Section~\ref{sec:probes}. Italic denotes full context; bold denotes the best
compressed method in each column.}
\label{tab:summary}
\begin{tabular}{lcccccccc}
\toprule
& \multicolumn{4}{c}{SGD $r{=}0.30$} & \multicolumn{4}{c}{MultiWOZ $r{=}0.30$} \\
\cmidrule(lr){2-5}\cmidrule(lr){6-9}
Method & P1$_\mathrm{s/l}$ & P2 & P3$_\mathrm{l}$ & P3$_\mathrm{s}$ & P1$_\mathrm{s/l}$ & P2 & P3$_\mathrm{l}$ & P3$_\mathrm{s}$ \\
\midrule
full context & $\mathit{0.487\,/\,0.850}$  & $\mathit{0.691}$  & $\mathit{0.728}$  & $\mathit{0.652}$  & $\mathit{0.430\,/\,0.867}$  & $\mathit{0.781}$  & $\mathit{0.505}$  & $\mathit{0.475}$ \\
\midrule
recency               & $0.010\,/\,0.135$ & $0.254$ & $0.067$ & $0.065$ & $0.002\,/\,0.092$ & $0.125$ & $0.040$ & $0.040$ \\
random selection        & $0.145\,/\,0.530$ & $0.278$ & $0.238$ & $0.162$ & $0.147\,/\,0.543$ & $\mathbf{0.314}$ & $0.145$ & $0.128$ \\
earliest turns     & $0.480\,/\,0.877$ & $0.156$ & $0.313$ & $0.160$ & $\mathbf{0.410}\,/\,\mathbf{0.867}$ & $0.296$ & $0.058$ & $0.048$ \\
uniform stride       & $\mathbf{0.493}\,/\,0.875$ & $0.283$ & $\mathbf{0.432}$ & $\mathbf{0.288}$ & $0.407\,/\,0.852$ & $0.286$ & $\mathbf{0.220}$ & $\mathbf{0.178}$ \\
attention/H2O proxy   & $\mathbf{0.493}\,/\,\mathbf{0.882}$ & $0.114$ & $0.288$ & $0.168$ & $0.408\,/\,0.858$ & $0.244$ & $0.068$ & $0.048$ \\
embedding MMR & $0.178\,/\,0.467$ & $\mathbf{0.309}$ & $0.185$ & $0.128$ & $0.128\,/\,0.482$ & $0.292$ & $0.120$ & $0.110$ \\
LLMLingua-2 & $0.193\,/\,0.502$ & $0.164$ & $0.117$ & $0.060$ & $0.117\,/\,0.522$ & $0.306$ & $0.117$ & $0.110$ \\
\bottomrule
\end{tabular}
\end{table}
\FloatBarrier

\begin{figure}[!htbp]
\centering
\begin{tikzpicture}[x=1mm,y=1mm,font=\footnotesize,
  posl/.style={line width=.9pt},cont/.style={line width=.9pt,dashed}]
\def\panel#1#2{%
 \begin{scope}[shift={(#1,0)}]
  \node[font=\small\bfseries] at (25,8) {#2};
  \foreach \r in {1,...,7}{\draw[black!18,line width=.3pt] (0,-6*\r+6)--(50,-6*\r+6);
    \node[anchor=east,black!60] at (-1.6,-6*\r+6) {\r};}
  \foreach \x/\l in {0/P1-strict,25/P1-loose,50/P2-strict}
    \node[anchor=north,black!70] at (\x,-42.5) {\l};
 \end{scope}}
\def\ln#1#2#3#4#5#6{\begin{scope}[shift={(#1,0)}]
  \draw[#2,#3] (0,-6*#4+6)--(25,-6*#5+6)--(50,-6*#6+6);
  \fill[#3] (0,-6*#4+6) circle(.8); \fill[#3] (25,-6*#5+6) circle(.8);
  \fill[#3] (50,-6*#6+6) circle(.8);\end{scope}}
\panel{0}{SGD}\panel{68}{MultiWOZ}
\ln{0}{posl}{black!55}{7}{7}{4}\ln{0}{posl}{blue!60!black}{6}{4}{3}
\ln{0}{posl}{orange!80!black}{3}{2}{6}\ln{0}{posl}{teal!70!black}{1.5}{3}{2}
\ln{0}{cont}{purple!60}{1.5}{1}{7}\ln{0}{cont}{cyan!60!black}{5}{6}{1}
\ln{0}{cont}{olive!80!black}{4}{5}{5}
\ln{68}{posl}{black!55}{7}{7}{7}\ln{68}{posl}{blue!60!black}{4}{4}{1}
\ln{68}{posl}{orange!80!black}{1}{1}{3}\ln{68}{posl}{teal!70!black}{3}{3}{5}
\ln{68}{cont}{purple!60}{2}{2}{6}\ln{68}{cont}{cyan!60!black}{5}{6}{4}
\ln{68}{cont}{olive!80!black}{6}{5}{2}
\begin{scope}[shift={(0,-53)}]
 \draw[posl,black!55](0,0)--(7,0);              \node[anchor=west] at (8,0) {recency};
 \draw[posl,blue!60!black](40,0)--(47,0);       \node[anchor=west] at (48,0) {random selection};
 \draw[posl,orange!80!black](80,0)--(87,0);     \node[anchor=west] at (88,0) {earliest turns};
 \draw[posl,teal!70!black](0,-6)--(7,-6);       \node[anchor=west] at (8,-6) {uniform stride};
 \draw[cont,purple!60](40,-6)--(47,-6);         \node[anchor=west] at (48,-6) {attention/H2O proxy};
 \draw[cont,cyan!60!black](80,-6)--(87,-6);     \node[anchor=west] at (88,-6) {embedding MMR};
 \draw[cont,olive!80!black](0,-12)--(7,-12);    \node[anchor=west] at (8,-12) {LLMLingua-2};
\end{scope}
\end{tikzpicture}
\caption{Rank changes among the seven compressed turn selectors at $r{=}0.30$.
Rank 1 is best and ties take the average rank. Each line follows one method
from P1-strict through P1-loose to P2-strict. Solid lines are the four
selectors that use position alone and dashed lines are the three that use
dialogue content, as defined in Section~\ref{sec:setup}. A line that falls
steeply between P1-loose and P2-strict marks a method that keeps the opening
but loses the revision. Table~\ref{tab:summary} reports the corresponding
accuracies.}
\label{fig:p1_p2_scatter}
\end{figure}
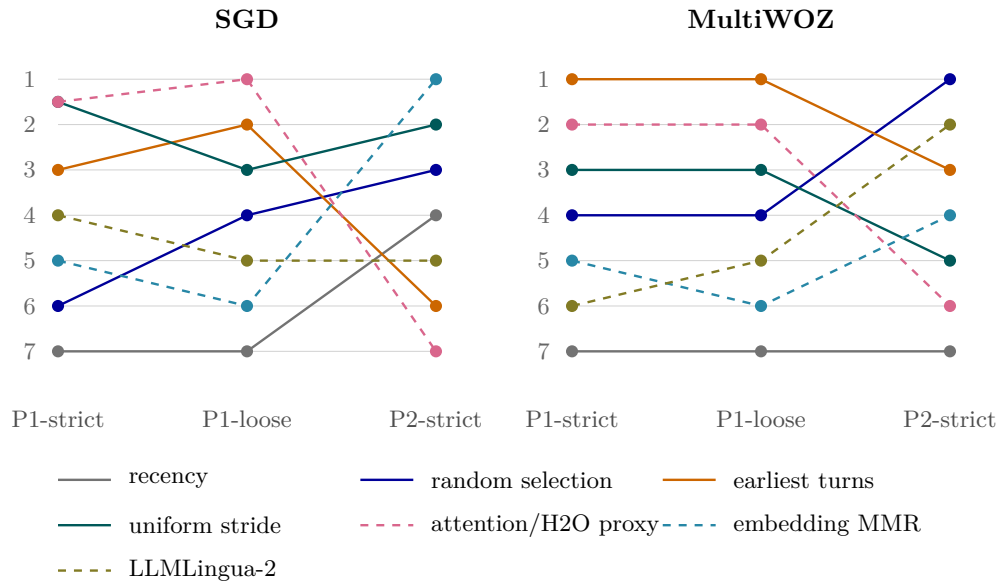

Figure~\ref{fig:p1_p2_scatter} compares the seven turn selectors at $r{=}0.30$. On SGD, the H2O proxy and uniform stride tie for the highest P1-strict score ($0.493$), whereas embedding MMR leads P2-strict ($0.309$) and scores $0.178$ on P1-strict. On MultiWOZ, earliest turns, the H2O proxy, and uniform stride fall within $0.004$ of each other on P1-strict ($0.407$--$0.410$), and random selection leads P2-strict ($0.314$). Across the seven selectors the Spearman rank correlation $\rho$ between
P1-strict and P2-strict is $-0.34$ on SGD and $+0.11$ on MultiWOZ. The
association is negative on SGD and weakly positive on MultiWOZ. On both datasets a different selector has the highest mean in each probe.
These comparisons describe target-specific performance on the sampled probe
sets, rather than isolating the effect of changing only the question.
Section~\ref{sec:retain_vs_read} separates answer retention from reader
success conditional on retention.

\paragraph{P1 scoring sensitivity}
Full context attains P1-strict $0.487$ on SGD and $0.430$ on MultiWOZ;
the loose scores are $0.850$ and $0.867$. These thresholds measure lexical
recovery at different levels of detail.

\paragraph{Aggregation can hide different strengths}
On SGD, earliest turns scores $0.480$ on P1-strict and $0.156$ on P2-strict,
while embedding MMR scores $0.178$ and $0.309$. One selector favors the initial
goal, the other the current value. A weighted average changes their order
with the probe mix.

\subsubsection{AI-assisted semantic evaluation of P1}
\label{sec:p1_semantic}

The lexical P1 rule rejects valid paraphrases, so this evaluation asks whether
it understates methods whose answers restate the goal in other words.

We evaluate $8{,}400$ Llama P1 answers at $r{=}0.30$: two datasets,
three seeds, and seven conditions including full context. The $3{,}049$
distinct reference--answer pairs each have one cached judgment. The judged set
covers full context and six turn selectors. Inputs contain the initial-goal
reference and reader answer, without method identities or expected rankings.

The judge is \texttt{gpt-5.4-mini-2026-03-17}, with temperature $0$, reasoning
effort \texttt{none}, and a $512$-token output limit. Full equivalence accepts
paraphrases and requires the action, target, and explicit task-relevant
constraints to retain their meaning. For example, ``find a restaurant'' omits
constraints from a request for a moderately priced restaurant in the west.
The judge records a short explanation and ambiguous decisions.

\paragraph{Agreement with external human judgments}
We assess the equivalence criterion on Answer Equivalence
(AE)~\citep{bulian2022answerequivalence}, which supplies human ratings of
answers to factual questions. Fixed, label-independent samples contain $200$
development and $600$ test pairs, with disjoint question IDs across splits.
Multiple human ratings are combined by majority vote; one development pair
and five test pairs have tied votes and remain unresolved. Human labels are
excluded from the judge input. The equivalence prompt and model are identical
for development, test, and TPBench. AE inputs include the question
and source passage; P1 inputs supply the initial-goal reference without the
remaining dialogue.

Development agreement is $0.859$ on $199$ resolved pairs. On the $595$
resolved test pairs, agreement is $0.837$, with a question-clustered bootstrap
$95\%$ interval of $[0.807,0.866]$ ($10{,}000$ resamples). Balanced accuracy
is $0.841$. The confusion counts are $201$ true positives, $297$ true
negatives, $64$ false positives, and $33$ false negatives, with no uncertain
model decisions. A fixed token-$F_1\geq0.5$ baseline attains accuracy $0.770$
and balanced accuracy $0.746$ on the same pairs. The paired accuracy gain is
$6.7$ percentage points, with a $95\%$ interval of $[3.0,10.4]$. All $600$
sampled test pairs differ under normalized exact match; that baseline always
predicts non-equivalence and scores $0.607$ accuracy. AE evaluates the full
equivalence decision in factual QA.

\paragraph{P1 semantic results}
Full context scores $0.482$ on SGD and $0.180$ on MultiWOZ. The MultiWOZ score
is an empirical uncompressed baseline, not an upper bound: the judge accepts
fewer than one fifth of the full-context answers as fully equivalent to the
initial reference. The reader requests a short noun phrase, whereas the judge
requires explicit task-relevant constraints; this difference can penalize
omissions. The low score limits absolute initial-goal recovery under this
reader/reference/judge setup even before compression. Source-dialogue-
clustered $95\%$ bootstrap intervals are $[0.433,0.533]$ and $[0.142,0.221]$
($5{,}000$ resamples). Repeated appearances across seeds remain in the same
cluster. Uncertain judgments remain in the denominator and receive no credit.

\begin{table}[!htbp]
\centering\footnotesize\setlength{\tabcolsep}{3.5pt}
\caption{AI-assisted full semantic equivalence of initial-goal answers at $r{=}0.30$,
averaged over three seeds ($200$ examples per dataset and seed). This analysis
covers six selectors and full context. LLMLingua-2 is outside the judged set. Italic denotes full context; bold denotes the best compressed method in each column.}
\label{tab:p1_semantic}
\begin{tabular}{lcc}
\toprule
Method & SGD & MultiWOZ \\
\midrule
full context (uncompressed) & $\mathit{0.482}$  & $\mathit{0.180}$ \\
\midrule
recency & $0.018$ & $0.007$ \\
random selection & $0.232$ & $0.082$ \\
earliest turns & $0.472$ & $\mathbf{0.172}$ \\
uniform stride & $0.458$ & $0.168$ \\
attention/H2O proxy & $\mathbf{0.480}$ & $0.162$ \\
embedding MMR & $0.187$ & $0.067$ \\
\bottomrule
\end{tabular}
\end{table}

In Table~\ref{tab:p1_semantic} the H2O proxy leads compressed-method
equivalence on SGD, and earliest turns leads on MultiWOZ. Correlations with P2-strict among these six selectors are $-0.60$ and
$+0.37$, respectively. The best selector differs between initial-goal and current-value recovery on both datasets.

\subsubsection{What a single average hides}
\label{sec:rank_inconsistency_example}

Let $R_1(C)$ and $R_2(C)$ denote the P1 and P2 accuracies of compressor $C$.
A weighted average $\alpha R_1(C)+(1-\alpha)R_2(C)$ depends on the probe mix
through $\alpha$. Two methods can have nearly equal averages even when one
leads on P1 and the other leads on P2.

\paragraph{Concrete instance}
At SGD $r{=}0.30$ (Table~\ref{tab:summary}), earliest turns scores
$0.480$ on P1-strict and $0.156$ on P2-strict, while embedding MMR scores
$0.178$ and $0.309$. Their weighted-score difference is
$0.302\alpha-0.153(1-\alpha)$, which changes sign at approximately
$\alpha=0.336$. Thus the aggregate ranking depends on the evaluation mix,
while the separate scores identify each selector's strength.

The P2 order could come from what each method keeps or from how the reader
reads what remains. The next three sections separate the two.

\subsubsection{Retention of the answer text}
\label{sec:gold_in_ctx}

This diagnostic checks whether a normalized current value appears as a complete
phrase in the compressed dialogue before the reader answers. It uses the same
phrase matcher as P2-loose and therefore separates text retention from reader
accuracy.

The probe builder uses normalized substring matching, whereas this diagnostic
uses phrase boundaries. Two MultiWOZ items contain only plural forms
(``colleges'' and ``parks'') for singular references (``college'' and ``park''),
accounting for the $159/161$ retention rate.

\begin{table}[!htbp]
\centering\footnotesize\setlength{\tabcolsep}{3.5pt}
\caption[P2 answer retention.]{Fraction of MultiWOZ P2 examples whose current
value remains in the compressed dialogue ($r{=}0.30$, $n{=}161$, seed $42$). Italic denotes full context; bold denotes the best compressed method in each column.}
\label{tab:gold_in_ctx}
\begin{tabular}{lc}
\toprule
Method & Answer retained \\
\midrule
full context (uncompressed) & $\mathit{0.988}$ \\
\midrule
LLMLingua-2 & $\mathbf{0.509}$ \\
embedding MMR                  & $0.422$ \\
attention/H2O proxy            & $0.429$ \\
uniform stride                & $0.472$ \\
earliest turns                       & $0.422$ \\
random selection                 & $0.491$ \\
recency                        & $0.205$ \\
\bottomrule
\end{tabular}
\end{table}

\subsubsection{Retention and reading}
\label{sec:retain_vs_read}

A method can lose a current value in two ways. The compressed dialogue may no
longer contain the value. Or it may contain the value and the reader may still
answer with something else. Table~\ref{tab:gold_in_ctx} measures the first way
on one cell. This section joins the two measurements example by example, on
both datasets and all three seeds.

We reuse the phrase matcher of Section~\ref{sec:gold_in_ctx}. For each example
we record whether the current value survives compression and whether the reader
answers correctly. Table~\ref{tab:retain_vs_read} reports the retention rate and
the accuracy on the retained examples. That accuracy is not a method ranking,
because each method is scored on the subset it keeps.

Retention and reader success both vary across methods. Conditional accuracy
on examples retaining the value ranges from $0.64$ to $0.83$ on SGD and
$0.57$ to $0.70$ on MultiWOZ. The P2-strict spreads across methods are $0.195$
and $0.189$, respectively.

Full context reads the retained value at $0.691$ on SGD and $0.790$ on
MultiWOZ. On MultiWOZ every compressed method reads it less often than full
context. On SGD four of the seven read it more often.

On SGD, uniform stride has the highest retention ($0.402$), whereas embedding
MMR has the highest P2-strict accuracy ($0.309$). Their conditional accuracies
are $0.664$ and $0.808$, respectively. LLMLingua-2
retains the value in $0.182$ of SGD examples and answers $0.827$ of that retained
subset correctly. Retention alone therefore does not determine the exact
method order.

\begin{table}[!htbp]
\centering\footnotesize\setlength{\tabcolsep}{3.5pt}
\caption[Retention and reading.]{P2-strict, answer retention, and accuracy on
the retained examples at $r{=}0.30$, averaged over seeds $\{42,43,44\}$. An
answer is retained when the complete current value occurs at phrase boundaries
in the compressed dialogue, and ``Read $|$ kept'' scores only those examples. Italic denotes full context; bold denotes the highest
compressed value in the first two columns of each corpus.}
\label{tab:retain_vs_read}
\begin{tabular}{lcccccc}
\toprule
& \multicolumn{3}{c}{SGD} & \multicolumn{3}{c}{MultiWOZ} \\
\cmidrule(lr){2-4}\cmidrule(lr){5-7}
Method & P2-strict & Retained & Read $|$ kept & P2-strict & Retained & Read $|$ kept \\
\midrule
full context (uncompressed) & $\mathit{0.691}$  & $\mathit{1.000}$  & $\mathit{0.691}$  & $\mathit{0.781}$  & $\mathit{0.988}$  & $\mathit{0.790}$ \\
\midrule
recency & $0.254$ & $0.309$ & $0.781$ & $0.125$ & $0.217$ & $0.573$ \\
random selection & $0.278$ & $0.372$ & $0.744$ & $\mathbf{0.314}$ & $\mathbf{0.505}$ & $0.613$ \\
earliest turns & $0.156$ & $0.211$ & $0.667$ & $0.296$ & $0.411$ & $0.696$ \\
uniform stride & $0.283$ & $\mathbf{0.402}$ & $0.664$ & $0.286$ & $0.481$ & $0.584$ \\
attention/H2O proxy & $0.114$ & $0.159$ & $0.645$ & $0.244$ & $0.403$ & $0.590$ \\
embedding MMR & $\mathbf{0.309}$ & $0.351$ & $0.808$ & $0.292$ & $0.421$ & $0.667$ \\
LLMLingua-2 & $0.164$ & $0.182$ & $0.827$ & $0.306$ & $0.446$ & $0.667$ \\
\bottomrule
\end{tabular}
\end{table}

\subsubsection{Results by slot type}
\label{sec:slot_retention}

This section asks whether the strongest P2 selector is the same for each
kind of revised slot.

Table~\ref{tab:slot_retention} groups the SGD examples by what the revised
slot holds. Random selection has the highest mean on time slots ($0.269$),
uniform stride on date slots ($0.387$), and embedding MMR on place/name slots
($0.417$) and the remaining slots ($0.372$). Thus the aggregate MMR lead does
not hold for every slot category.

LLMLingua-2 retains $0.208$ of time values and
$0.149$ of date values, compared with $0.180$ on place/name and $0.191$ on
other slots. Its retention is therefore close to even across slot types.

\begin{table}[!htbp]
\centering\footnotesize\setlength{\tabcolsep}{3.5pt}
\caption[Retention by slot type.]{SGD P2-strict and answer retention by slot
type at $r{=}0.30$, averaged over seeds $\{42,43,44\}$. Each entry reads
P2-strict~/~retention, and the four groups together cover every example. Bold
denotes the highest P2-strict in each column.}
\label{tab:slot_retention}
\begin{tabular}{lcccc}
\toprule
Method & Time & Date & Place and name & Other \\
\midrule
recency & $0.255\,/\,0.347$ & $0.216\,/\,0.299$ & $0.252\,/\,0.243$ & $0.316\,/\,0.331$ \\
random selection & $\mathbf{0.269}\,/\,0.346$ & $0.239\,/\,0.303$ & $0.340\,/\,0.396$ & $0.273\,/\,0.489$ \\
earliest turns & $0.100\,/\,0.092$ & $0.157\,/\,0.185$ & $0.269\,/\,0.332$ & $0.128\,/\,0.327$ \\
uniform stride & $0.231\,/\,0.390$ & $\mathbf{0.387}\,/\,0.453$ & $0.279\,/\,0.358$ & $0.228\,/\,0.398$ \\
attention/H2O proxy & $0.072\,/\,0.063$ & $0.141\,/\,0.157$ & $0.152\,/\,0.234$ & $0.116\,/\,0.255$ \\
embedding MMR & $0.219\,/\,0.269$ & $0.299\,/\,0.307$ & $\mathbf{0.417}\,/\,0.482$ & $\mathbf{0.372}\,/\,0.430$ \\
LLMLingua-2 & $0.194\,/\,0.208$ & $0.143\,/\,0.149$ & $0.168\,/\,0.180$ & $0.148\,/\,0.191$ \\
\bottomrule
\end{tabular}
\end{table}

\subsection{The P3 gap to full context}
\label{sec:f2}

P3 evaluates whether a compressed context preserves both the initial goal and
a late slot update. Every compressed method scores below full context on it,
and this section reports by how much and tests three ways the gap could be an
artifact of the sample or the scorer.

Uniform stride is the
strongest compressed turn selector on both datasets. It also leads in each of
the three seeds. Its loose score is $29.7$ percentage points below full context
on SGD and $28.5$ points below on MultiWOZ. The strict gaps are $36.3$ and
$29.7$ points.

Embedding MMR leads SGD P2. Random selection leads MultiWOZ P2. Neither method
leads P3. Current-value recovery alone does not identify the best method when
the reader also needs the initial goal. A matched first-user-plus-recent baseline also leaves a substantial P3 gap to full context (Section~\ref{sec:first_recent}).

\subsubsection{A matched first-user-plus-recent baseline}
\label{sec:first_recent}

This baseline tests whether anchoring the first user turn and reserving the
remaining budget for recent dialogue is sufficient for joint recovery. A late
P3 update is constrained to the second half, but need not be near the end.

The matched comparison fixes the first USER turn and allocates the remaining
turn budget to the most recent turns. All methods use the same probe examples
and Llama reader, with greedy decoding, $96$ output tokens and eager attention
for P1/P2, and $128$ output tokens with SDPA for P3. The model revision is
\texttt{0e9e39f249a16976\allowbreak 918f6564b8830bc894c89659}; inference uses bfloat16,
PyTorch 2.5.1, Transformers 5.13.1, and batches of up to eight requests.

\begin{table}[!htbp]
\centering\footnotesize\setlength{\tabcolsep}{3.5pt}
\caption{Matched first-user-plus-recent comparison at $r{=}0.30$, three seeds,
scored with the segment-local P3-strict rule. This experiment uses the same
$200$ examples per seed as Table~\ref{tab:summary} but a separate generation
run. Its full-context values differ from that table by at most $0.007$. Italic denotes full context; bold denotes the best compressed method in each column.}
\label{tab:first_recent}
\begin{tabular}{lcc}
\toprule
Method & SGD P3 & MultiWOZ P3 \\
\midrule
full context & $\mathit{0.645}$  & $\mathit{0.478}$ \\
\midrule
earliest turns & $0.162$ & $0.048$ \\
recency & $0.065$ & $0.040$ \\
first USER + recent & $0.247$ & $0.137$ \\
uniform stride & $\mathbf{0.287}$ & $\mathbf{0.178}$ \\
\bottomrule
\end{tabular}
\end{table}

In Table~\ref{tab:first_recent} the first-user-plus-recent P3-strict
differences from full context are
$-39.8$ points on SGD ($95\%$ CI $[-46.4,-33.1]$) and $-34.2$ points on
MultiWOZ ($[-41.4,-27.2]$), using $5{,}000$ source-dialogue-clustered
resamples. It improves on earliest turns and recency, but retaining both ends
of the dialogue leaves a large joint-recovery gap. Its differences from
uniform stride have intervals containing zero.

The same selector attains P1-strict $0.488$ and P2-strict $0.163$ on SGD,
and P1-strict $0.443$ and P2-strict $0.088$ on MultiWOZ. These scores describe
its within-probe behavior; the two metrics have different scales.

\subsubsection{Sensitivity to excluding alias-equivalent updates}
\label{sec:p3_alias_sensitivity}

The primary P3 pool retains late annotation changes, including pairs that
P2's alias heuristic would exclude. Applying that same heuristic identifies
$48$, $53$, and $53$ of the $200$ SGD examples at seeds $42$, $43$, and $44$,
and $19$, $17$, and $20$ of the MultiWOZ examples. These are heuristic matches,
not human judgments that every pair is semantically identical. Examples include
``afternoon 12:45'' versus ``12:45 PM'' and ``Vancouver, BC'' versus
``Vancouver.''

Table~\ref{tab:p3_alias_sensitivity} excludes these pairs from the saved
segment-local predictions. We do not generate new answers or replenish the
samples to $200$. Each number is the arithmetic mean of three seed accuracies
with the reduced denominators. The full-context advantage and the identities of the highest-mean selectors
remain in these reduced sets. Two selectors fall much further than the rest on
SGD P3-strict: earliest turns drops from $0.160$ to $0.036$ and the H2O proxy
from $0.168$ to $0.038$. Their scores on the primary pool come mostly from the
excluded pairs, so their position in the middle of the SGD ranking depends on
which pairs the pool keeps. The leading selector and the full-context
advantage hold in both pools.

\begin{table}[!htbp]
\centering\footnotesize\setlength{\tabcolsep}{3.5pt}
\caption{Saved-output P3 sensitivity after excluding pairs accepted by the
P2 alias heuristic. SGD denominators are $152/147/147$ and MultiWOZ
$181/183/180$ for seeds $42/43/44$. U denotes uniform stride and L the
LLMLingua-2. ``Best'' is the highest mean among seven turn
selectors, selected separately for each threshold.}
\label{tab:p3_alias_sensitivity}
\begin{tabular}{lcccc}
\toprule
& \multicolumn{2}{c}{P3-loose} & \multicolumn{2}{c}{P3-strict} \\
\cmidrule(lr){2-3}\cmidrule(lr){4-5}
Reader / corpus & Full & Best & Full & Best \\
\midrule
Llama SGD & $0.700$ & $0.437$ (U) & $0.601$ & $0.262$ (U) \\
Llama MultiWOZ & $0.518$ & $0.222$ (U) & $0.502$ & $0.189$ (U) \\
Mistral SGD & $0.310$ & $0.256$ (U) & $0.260$ & $0.143$ (U) \\
Mistral MultiWOZ & $0.165$ & $0.086$ (L) & $0.151$ & $0.083$ (L) \\
\bottomrule
\end{tabular}
\end{table}

For the saved ChunkKV seed-42 predictions, the same exclusion leaves $152$
SGD and $181$ MultiWOZ examples. Loose/strict scores are $0.171/0.151$ and
$0.105/0.088$, respectively. These are single-seed diagnostics.

\subsubsection{Sensitivity to scoring the whole answer field}
\label{sec:p3_segment_sensitivity}

This check asks how much the P3 results depend on scoring the goal and the
value in separate segments.

The reported P3 scorer parses the answer into a goal segment and a
\texttt{slot: value} segment. A deterministic parser accepts one semicolon
separating a nonempty goal from a nonempty \texttt{slot: value} expression. The
first colon in the second part separates the label from the value, and
subsequent colons, including those in times, are preserved. An answer the
parser cannot read scores zero. The alternative is to apply both criteria to
the complete answer field. That rule credits an answer whose value words appear
in the goal part, or the reverse, so it can pass an answer that places the
right content in the wrong place.

This comparison covers all $19{,}600$ saved P3 answers: $9{,}600$ each for
Llama and Mistral and $400$ for ChunkKV. Whole-field scoring passes $1{,}896$
Llama answers, of which $203$ fail under segment-local scoring: $41$ because
the parser cannot read them and $162$ because a match lands outside its
segment. Mistral passes $1{,}269$ and loses $424$, split $273$ and $151$.
ChunkKV passes $52$ and loses $7$, split $1$ and $6$.

\begin{table}[!htbp]
\centering\footnotesize\setlength{\tabcolsep}{3.5pt}
\caption[Whole-field and segment-local P3 scoring.]{P3-strict on Llama under
the two scorers, averaged over seeds $\{42,43,44\}$ ($200$ examples per dataset
and seed). ``Segment'' is the scorer used in the main tables and ``Whole''
applies both criteria to the complete answer. ``Parsed'' is the share of
answers the segment parser reads, and ``Seg.\,$|$\,parsed'' scores only those.
Italic denotes full context; bold denotes the best compressed method in each
accuracy column.}
\label{tab:p3_segment_sensitivity}
\begin{tabular}{lcccccccc}
\toprule
& \multicolumn{4}{c}{SGD} & \multicolumn{4}{c}{MultiWOZ} \\
\cmidrule(lr){2-5}\cmidrule(lr){6-9}
Method & Segment & Whole & Parsed & Seg.\,$|$\,parsed & Segment & Whole & Parsed & Seg.\,$|$\,parsed \\
\midrule
full context & $\mathit{0.652}$  & $\mathit{0.658}$  & $\mathit{0.977}$  & $\mathit{0.667}$  & $\mathit{0.475}$  & $\mathit{0.497}$  & $\mathit{0.945}$  & $\mathit{0.503}$ \\
\midrule
recency             & $0.065$ & $0.075$ & $0.870$ & $0.075$ & $0.040$ & $0.058$ & $0.885$ & $0.045$ \\
random selection    & $0.162$ & $0.182$ & $0.967$ & $0.167$ & $0.128$ & $0.148$ & $0.907$ & $0.142$ \\
earliest turns      & $0.160$ & $0.172$ & $0.953$ & $0.168$ & $0.048$ & $0.085$ & $0.835$ & $0.058$ \\
uniform stride      & $\mathbf{0.288}$ & $\mathbf{0.298}$ & $0.973$ & $\mathbf{0.296}$ & $\mathbf{0.178}$ & $\mathbf{0.222}$ & $0.967$ & $\mathbf{0.184}$ \\
attention/H2O proxy & $0.168$ & $0.183$ & $0.908$ & $0.185$ & $0.048$ & $0.097$ & $0.725$ & $0.066$ \\
embedding MMR       & $0.128$ & $0.157$ & $0.762$ & $0.169$ & $0.110$ & $0.110$ & $0.750$ & $0.147$ \\
LLMLingua-2 & $0.060$ & $0.072$ & $0.860$ & $0.070$ & $0.110$ & $0.147$ & $0.805$ & $0.137$ \\
\bottomrule
\end{tabular}
\end{table}

Whole-field scoring never lowers a Llama cell, and it raises all but one:
MultiWOZ embedding MMR scores $0.110$ under both. The SGD order is unchanged.
On MultiWOZ, some relative orders and ties change. Uniform stride leads the compressed methods on both
datasets under both scorers.

The full-context advantage over the strongest compressed method is $28.5$ to
$36.3$ points across the two datasets and thresholds. We bootstrap that gap
over source dialogues ($5{,}000$ paired resamples). On SGD the interval is
$[23.5,35.9]$ points under the loose threshold and $[30.1,42.5]$ under the
strict one. On MultiWOZ the two intervals are $[21.3,35.7]$ and $[22.4,36.9]$.
None of the four contains zero.

Segment-local scoring charges for response format, because it gives zero to an
answer the parser cannot read. The share of answers that parse ranges from
$0.725$ to $0.977$ across methods on Llama. On Mistral it ranges from $0.583$
to $0.852$, and full context has the lowest share of any condition. The last
column of Table~\ref{tab:p3_segment_sensitivity} scores only the answers that
parse. The full-context advantage holds in that column on
both datasets, and so does the lead of uniform stride among compressed methods.
The whole-field rule can credit an answer that puts the right content in the
wrong segment. All three scoring views preserve the Llama full-context
advantage and the uniform-stride lead shown here.

On Mistral the two scorers disagree about the leading selector only for
MultiWOZ-loose: whole-field scoring favors uniform stride, whereas
segment-local scoring favors LLMLingua-2. Uniform stride
leads both SGD thresholds under both scorers, and LLMLingua-2 leads
MultiWOZ-strict under both. Full context remains above the
strongest compressed mean by $6.3$ to $13.0$ points under segment-local scoring.

\subsection{Deleting the turn that carries the update}
\label{sec:finding3}
\label{sec:f3}

A low score shows that a method lost something. This section identifies which
turns the reader depends on, by deleting one turn at a time from the
uncompressed dialogue and watching the answer change.

The evidence analysis first locates the final value throughout each dialogue. The update turn itself does not
contain the complete normalized reference in $68/128$ SGD examples and
$34/161$ MultiWOZ examples. The union of the update turn and answer-bearing turns contains more than one
turn in $106/128$ SGD examples and $120/161$ MultiWOZ examples. A system turn contains
the answer in $102/128$ and $108/161$ examples. These system turns often
confirm or restate the value after the user update. Evidence for the final value is therefore distributed across update and answer-bearing turns.

Two paired comparisons measure the contribution of these turns. The first removes the update turn and
compares it with one matched irrelevant turn. The second removes the update
turn and every answer-bearing turn. Its control removes the same number of
irrelevant turns with the same speaker composition. These controls measure the
effect of deleting evidence separately from the effect of shortening context.

\begin{table}[!htbp]
\centering\footnotesize\setlength{\tabcolsep}{3.5pt}
\caption{Strict P2 accuracy in the paired deletion experiment, on the seed-$42$
P2 sets. The single-turn comparison uses $128$ SGD and $161$ MultiWOZ examples.
The comparison that removes the update turn together with all answer-bearing
turns uses $128$ and $159$, respectively. The displayed full-context row uses
the single-turn sample; all-evidence comparisons pair their last two rows.}
\label{tab:cf_ablation_main}
\begin{tabular}{lcc}
\toprule
Condition & SGD & MultiWOZ \\
\midrule
full context & $\mathit{0.672}$  & $\mathit{0.801}$ \\
\midrule
remove update turn & $0.508$ & $0.391$ \\
remove matched single turn & $0.672$ & $0.801$ \\
remove update turn and all answer-bearing turns & $0.070$ & $0.019$ \\
remove matched control turns & $0.648$ & $0.811$ \\
\bottomrule
\end{tabular}
\end{table}

Table~\ref{tab:cf_ablation_main} shows that removing only the update turn
lowers strict accuracy on both datasets. The
single-turn controls leave accuracy at the full-context level. The control
prompt differs from full context in every example, because one turn is removed
from each. On SGD the reader nevertheless gives the same answer on all $128$
examples. On MultiWOZ four answers change and the two totals coincide. Removing the update
turn together with all answer-bearing turns produces a much larger decrease. The matched controls do
not show that decrease. Exact McNemar tests compare each evidence deletion with its matched control
among items answered correctly in full context. All four tests give
$p\leq2.4\!\times\!10^{-7}$.

The single-turn intervention reduces accuracy by $41.0$ points on MultiWOZ
and $16.4$ on SGD, demonstrating the reader's dependence on the annotated
update even when other turns retain answer evidence.

\subsubsection{Locating update turns and testing their removal}
\label{sec:p2_deletion}

The update-turn mapping replays the SGD and MultiWOZ conversion with its
recorded seed and iteration order. A question maps to a user turn when its
slot, old value, new value, question, and answer match the stored row and that
turn contains the corresponding state change. All $5{,}227$ SGD and $3{,}885$
MultiWOZ source questions match. Each of the $949$ SGD candidates has one
possible update turn. In MultiWOZ, $722$ have one, $15$ have two, and one has
three because an identical state change is repeated. The recorded conversion
order selects one turn in these $16$ cases. Missing update turns, updates with no
corresponding dialogue turn, and selected system turns all have count zero.
Requiring the selected turn to be the final update to its slot leaves $791$ SGD
and $664$ MultiWOZ candidates.

The field \texttt{s\_origin} stores the update turn. The field
\texttt{s\_evidence} stores every user or system turn containing the complete
normalized reference at phrase boundaries. Their union is stored as
\texttt{s\_transition\_support}.

The update-turn control removes one non-answer turn with the same speaker and
the closest word length and position. The all-evidence control reuses that turn
and adds distinct non-answer turns to match the number and speaker composition
of the remaining answer-bearing turns; word length and position again break
ties. This construction succeeds for all $128$ SGD examples and $159/161$
MultiWOZ examples. The two unmatched MultiWOZ examples remain in the
update-turn comparison.

\begin{figure}[!htbp]
\centering
\begin{tikzpicture}[font=\footnotesize,
  col/.style={draw=black!55,rounded corners=2pt,inner sep=4pt,text width=36mm,align=center},
  head/.style={col,fill=black!8,font=\footnotesize\bfseries},
  body/.style={col,align=center},
  num/.style={col,align=center},
  ar/.style={-{Latex[length=1.8mm]},black!60}]
\node[head] (h1) at (0,0) {full context};
\node[head,right=5mm of h1] (h2) {remove the update turn};
\node[head,right=5mm of h2] (h3) {remove one matched\\ irrelevant turn};
\node[body,below=2mm of h1] (b1) {initial goal turn\\ unrelated turn\\ \textbf{update turn}\\ answer-repeating turn};
\node[body,below=2mm of h2,draw=black,very thick] (b2) {initial goal turn\\ unrelated turn\\ \textbf{\textcolor{red!70!black}{update turn removed}}\\ answer-repeating turn};
\node[body,below=2mm of h3] (b3) {initial goal turn\\ \textcolor{red!70!black}{unrelated turn removed}\\ \textbf{update turn}\\ answer-repeating turn};
\node[num,below=2mm of b1] (n1) {current value \textbf{0.801}\\ superseded value \textbf{0.062}\\ no answer \textbf{0.000}};
\node[num,below=2mm of b2,draw=black,very thick] (n2) {current value \textbf{0.391}\\ superseded value \textbf{0.286}\\ no answer \textbf{0.000}};
\node[num,below=2mm of b3] (n3) {current value \textbf{0.801}\\ superseded value \textbf{0.062}\\ no answer \textbf{0.000}};
\foreach \a/\b in {b1/n1,b2/n2,b3/n3} \draw[ar] (\a)--(\b);
\end{tikzpicture}
\caption{Deleting the update turn moves the reader from the current value to
the value it replaced, and the matched control does not. Each column is one
deletion condition, red marks the turn removed, and the heavy boxes mark the
condition whose answers change. Rates are MultiWOZ P2 at seed $42$
($n{=}161$) from Tables~\ref{tab:cf_ablation} and~\ref{tab:old_value}.}
\label{fig:deletion_mechanism}
\end{figure}
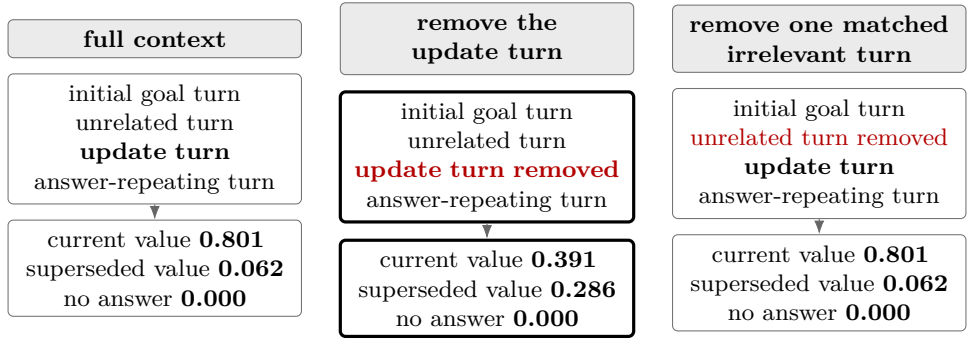

Figure~\ref{fig:deletion_mechanism} summarizes the three conditions and what
the reader answers in each. Every condition uses the same full-context P2
prompt and whole-turn deletion.
Llama-3.1-8B-Instruct runs with greedy decoding,
\texttt{max\_new\_tokens} $=128$, \texttt{max\_input\_tokens} $=7{,}168$, and
scaled dot-product attention (SDPA). Reader errors and malformed outputs count
as wrong.

\begin{table}[!htbp]
\centering\footnotesize\setlength{\tabcolsep}{3.5pt}
\caption{P2 deletion experiment on seed $42$, reported as strict~/~loose
accuracy. SGD uses $128$ examples throughout. MultiWOZ uses $161$ for the first
three rows and $159$ for the last two, because two items have no valid matched
control for the all-evidence comparison. The displayed full-context row uses
the single-turn sample.}
\label{tab:cf_ablation}
\begin{tabular}{lcc}
\toprule
Condition & SGD & MultiWOZ \\
\midrule
full context & $\mathit{0.672~/~0.695}$  & $\mathit{0.801~/~0.820}$ \\
\midrule
remove update turn              & $0.508~/~0.531$ & $0.391~/~0.422$ \\
remove matched single turn      & $0.672~/~0.695$ & $0.801~/~0.814$ \\
remove update turn and all answer-bearing turns & $0.070~/~0.078$ & $0.019~/~0.019$ \\
remove matched control turns    & $0.648~/~0.672$ & $0.811~/~0.830$ \\
\bottomrule
\end{tabular}
\end{table}

In Table~\ref{tab:cf_ablation}, among items answered correctly in full
context, update-turn deletion changes
$23/86$ SGD and $66/129$ MultiWOZ answers from correct to wrong, compared with
$0/86$ and $2/129$ for the controls. The exact paired McNemar $p$-values are
$2.38\times10^{-7}$ and $1.08\times10^{-19}$. Removing the update turn together
with all answer-bearing turns changes $78/86$ and $125/128$ answers, compared with $3/86$ and $2/128$ for
the controls; the corresponding $p$-values are $1.03\times10^{-21}$ and
$1.88\times10^{-37}$. These tests compare paired outcomes within the
full-context-correct subsets.

\paragraph{Comparison with the synthetic set}
The synthetic comparison evaluates full context, support deletion, and
matched control deletion on $100$ examples from the free-form set (Section~\ref{sec:fmts}) whose answers are confined
to three authored turns. Strict uses normalized exact match; loose accepts
phrase-bounded containment in either direction for these free-form answers.
Strict accuracy is $0.80$ in full context,
$0.00$ after removing those three turns, and $0.80$ after the matched control deletion;
loose accuracy is $0.97$, $0.00$, and $0.96$. The synthetic set confines answer
evidence to the three support turns, whereas SGD and MultiWOZ distribute it
across update turns and repeated mentions.

\paragraph{One phrase matcher for evidence and scoring}
The loose P2 scorer and \texttt{s\_evidence} call the same function: the
complete normalized reference must occur at phrase boundaries in the candidate
answer or turn. Validation recomputes the scores of $40{,}124$ saved slot-value
predictions and agrees with every stored score. The P3 scorer also applies
phrase-boundary checks when matching the complete reference value.

\subsubsection{What the reader answers when the evidence is gone}
\label{sec:old_value}

Table~\ref{tab:old_value} distinguishes explicit abstention from replies that
match the superseded value. It does not classify the mechanism behind every
other incorrect answer. Explicit no-answer
replies stay at or below $0.013$ in every condition, including the condition
in which strict accuracy falls to $0.019$. Replies matching the value the user replaced become more frequent. That rate rises from $0.094$ to $0.234$ on SGD and
from $0.062$ to $0.286$ on MultiWOZ when the update turn is removed, and to
$0.445$ and $0.484$ when the answer-bearing turns are removed as well. The
matched controls produce much smaller changes. Evidence deletion therefore
shifts responses toward superseded values while explicit abstention remains
rare.

\begin{table}[!htbp]
\centering\footnotesize\setlength{\tabcolsep}{3.5pt}
\caption{What the reader answers when evidence is deleted, on the seed-$42$ P2 sets. Abstention is the rate of explicit no-answer replies and old value uses
normalized exact matching against the replaced value. SGD uses $n=128$;
MultiWOZ uses $n=161$, and $n=159$ for the last two rows.}
\label{tab:old_value}
\begin{tabular}{lcccc}
\toprule
& \multicolumn{2}{c}{SGD} & \multicolumn{2}{c}{MultiWOZ} \\
\cmidrule(lr){2-3}\cmidrule(lr){4-5}
Condition & Abstain & Old value & Abstain & Old value \\
\midrule
full context & $\mathit{0.000}$  & $\mathit{0.094}$  & $\mathit{0.000}$  & $\mathit{0.062}$ \\
\midrule
$-$ update turn & $0.008$ & $0.234$ & $0.000$ & $0.286$ \\
\quad matched control & $0.000$ & $0.094$ & $0.000$ & $0.062$ \\
$-$ update and answer-bearing turns & $0.008$ & $0.445$ & $0.013$ & $0.484$ \\
\quad matched control & $0.000$ & $0.102$ & $0.000$ & $0.044$ \\
\bottomrule
\end{tabular}
\end{table}

\subsection{Wider budgets and cache-level compression}
\label{sec:f4}

The main tables keep $30\%$ of the turns and compress the prompt. This section
loosens the budget and then switches to methods that compress the reader's
attention cache instead of the text.

Full context improves P2-strict over the best compressed turn selector by
$38.2$ points on SGD and $46.7$ points on MultiWOZ. For prompt-side selection,
retaining $30\%$ of input tokens would reduce the dense prefill attention term
by about $11\times$ and input KV-cache memory by about $3.3\times$.
These scaling factors apply to token retention, and the experimental budget
$r$ counts turns. They are theoretical scaling factors rather than measured
end-to-end speedups.

SnapKV, PyramidKV, and StreamingLLM were tested at $r{=}0.30$ on P2. On MultiWOZ P2, SnapKV
scores $0.689$. PyramidKV scores $0.683$. The best turn selector on the same
seed scores $0.329$.

\subsubsection{Wider turn budgets}
\label{sec:wider_ratios}

These budgets ask whether the P2 gap to full context persists when a
compressed context may keep more turns.

The wider-budget evaluation covers full context and four turn selectors (recency, random
selection, earliest turns, and uniform stride) at
$r\in\{0.50,0.70\}$ on both datasets for P2. Results use seeds $\{42,43,44\}$, $200$ examples per seed
before P2 filtering, and the Llama reader with greedy decoding and $128$ output
tokens. Full context uses the same samples within each comparison.

\begin{table}[!htbp]
\centering\footnotesize\setlength{\tabcolsep}{3.5pt}
\caption[P2 at wider turn budgets.]{P2 strict~/~loose accuracy at
$r\in\{0.50,0.70\}$, mean $\pm$ standard deviation over three seeds. The
three samples contain $\{128,125,132\}$ SGD and $\{161,168,168\}$ MultiWOZ
examples. Full context does not use the budget, so its row repeats across the
two columns. Italic denotes full context; bold denotes the best compressed
method in each column within each corpus.}
\label{tab:P2_wider_ratios}
\begin{tabular}{@{}lcc@{}}
\toprule
Method & $r{=}0.50$ & $r{=}0.70$ \\
\midrule
\multicolumn{3}{@{}l}{\textbf{MultiWOZ}} \\
full context (uncompressed) & $\mathit{0.779 \pm 0.041}$ / $\mathit{0.805 \pm 0.027}$  & $\mathit{0.779 \pm 0.041}$ / $\mathit{0.805 \pm 0.027}$ \\
\midrule
recency & $0.348 \pm 0.052$ / $0.390 \pm 0.051$ & $0.583 \pm 0.021$ / $0.608 \pm 0.021$ \\
random selection & $0.493 \pm 0.011$ / $0.553 \pm 0.006$ & $0.607 \pm 0.042$ / $0.650 \pm 0.041$ \\
earliest turns & $0.527 \pm 0.005$ / $0.575 \pm 0.020$ & $\mathbf{0.741} \pm 0.035$ / $0.759 \pm 0.029$ \\
uniform stride & $\mathbf{0.575} \pm 0.046$ / $0.702 \pm 0.057$ & $0.640 \pm 0.029$ / $0.670 \pm 0.027$ \\
\addlinespace
\multicolumn{3}{@{}l}{\textbf{SGD}} \\
full context (uncompressed) & $\mathit{0.693 \pm 0.019}$ / $\mathit{0.714 \pm 0.017}$ & $\mathit{0.693 \pm 0.019}$ / $\mathit{0.714 \pm 0.017}$ \\
\midrule
recency & $0.467 \pm 0.024$ / $0.504 \pm 0.034$ & $\mathbf{0.644} \pm 0.014$ / $0.667 \pm 0.019$ \\
random selection & $\mathbf{0.476} \pm 0.042$ / $0.501 \pm 0.036$ & $0.543 \pm 0.029$ / $0.567 \pm 0.024$ \\
earliest turns & $0.363 \pm 0.058$ / $0.376 \pm 0.056$ & $0.587 \pm 0.015$ / $0.605 \pm 0.011$ \\
uniform stride & $0.354 \pm 0.021$ / $0.380 \pm 0.027$ & $0.532 \pm 0.058$ / $0.547 \pm 0.045$ \\
\bottomrule
\end{tabular}
\end{table}

\par\medskip\noindent
In Table~\ref{tab:P2_wider_ratios}, raising $r$ from $0.50$ to $0.70$
lifts P2-strict in all eight dataset--method comparisons, by $0.065$--$0.235$. On MultiWOZ at $r{=}0.70$,
earliest turns scores $0.741\pm0.035$ against $0.779\pm0.041$ for full context,
a $3.8$-point gap. The joint P3 probe is evaluated only at $r{=}0.30$.

\subsubsection{KV-cache compression results}
\label{sec:kv_methods}
\label{sec:chunkkv}

These methods compress the attention cache instead of selecting turns, so this
analysis asks whether the probe-dependent pattern also appears in a different
method family.

We test ChunkKV~\citep{liu2025chunkkv}, SnapKV~\citep{li2024snapkv},
PyramidKV~\citep{cai2025pyramidkvdynamickvcache}, and
StreamingLLM~\citep{xiao2024streamingllm} through the kvpress implementation
library\footnote{\url{https://github.com/NVIDIA/kvpress}} at seed $42$. All four
have P2 results at $r{=}0.30$. P3 is reported for ChunkKV.
ChunkKV uses \texttt{chunk\_length} $=20$, drop ratio $0.7$, window $64$, kernel
$5$, and context-only compression. Its P3 reader uses eager attention,
$128$ output tokens, and a $7{,}168$-token input limit.

\begin{table}[!htbp]
\centering\footnotesize\setlength{\tabcolsep}{3.5pt}
\caption[KV-cache results at $r{=}0.30$.]{Strict accuracy for KV-cache methods
at $r{=}0.30$, seed $42$. The P3 rows use the segment-local scorer. Higher is
better. Bold denotes the best KV-cache method in each row with more than one
entry.
The last column repeats the best turn selector on the same seed; MMR/random
denotes a tie between embedding MMR and random selection. A dash marks
a condition we did not evaluate.}
\label{tab:kv_r03}
\begin{tabular}{lccccc}
\toprule
Setting & ChunkKV & SnapKV & PyramidKV & StreamingLLM & Best turn selector \\
\midrule
SGD P2 $r{=}0.30$        & $0.133$ & $0.469$          & $\mathbf{0.492}$ & $0.297$ & $0.297$ (MMR/random)     \\
SGD P3-strict $r{=}0.30$ & $0.140$ & --- & --- & --- & $0.280$ (uniform stride) \\
MultiWOZ P2 $r{=}0.30$         & $0.093$ & $\mathbf{0.689}$ & $0.683$          & $0.584$ & $0.329$ (LLMLingua-2) \\
MultiWOZ P3-strict $r{=}0.30$ & $0.085$ & --- & --- & --- & $0.195$ (uniform stride) \\
\bottomrule
\end{tabular}
\end{table}

At $r{=}0.30$, SnapKV, PyramidKV, and StreamingLLM produce a reader error on
$0$ to $5$ of the reported P2 examples per dataset, and those rows are scored
as wrong. On MultiWOZ P2 they score $0.689$, $0.683$, and
$0.584$, compared with $0.329$ for the best turn selector on the same seed.
SnapKV and PyramidKV differ by at most $0.024$ across the two P2 settings.

The MultiWOZ P2 sensitivity sweep at $r{=}0.30$ compares
\texttt{chunk\_length} $\in\{10,20,40\}$ and yields strict scores of $0.087$,
$0.087$, and $0.050$. A chunk length of $40$ therefore costs about four points. Table~\ref{tab:kv_r03} reports the sweep run of the
\texttt{chunk\_length} $=20$ setting.

\subsubsection{ChunkKV chunk length}
\label{sec:chunklen}

ChunkKV prunes the cache in contiguous chunks, so the chunk length sets how
coarse the pruning is. Table~\ref{tab:chunklen} varies it on the MultiWOZ P2
set at $r{=}0.30$ ($n{=}161$). Chunk lengths of $10$ and $20$ give the same
strict score, and $40$ lowers it to $0.050$. The drop coincides with a higher malformed-output rate: it rises from $0.124$ at chunk length $10$ to $0.199$ at $20$ and
$0.478$ at $40$, while the abstention rate stays near $0.32$. The main
comparison uses chunk length $20$.

\begin{table}[!htbp]
\centering\footnotesize\setlength{\tabcolsep}{3.5pt}
\caption{ChunkKV chunk length on the MultiWOZ P2 set at $r{=}0.30$ ($n{=}161$,
seed $42$).}
\label{tab:chunklen}
\begin{tabular}{lcccc}
\toprule
Chunk length & P2-strict & P2-loose & Abstain & Malformed \\
\midrule
$10$ & $0.087$ & $0.099$ & $0.323$ & $0.124$ \\
$20$ & $0.087$ & $0.093$ & $0.342$ & $0.199$ \\
$40$ & $0.050$ & $0.056$ & $0.311$ & $0.478$ \\
\bottomrule
\end{tabular}
\end{table}

\subsection{Another reader, corpus, and language}
\label{sec:robustness}
\label{sec:f5}

The main results use one reader on two English corpora. This section repeats
the current-value and joint probes with a second reader, an English long-term
memory corpus, and a Chinese corpus, to see whether the same methods lead.
The reader comparison uses three sampled probe sets. For LongMemEval-KU and
RiSAWOZ, we evaluate the complete eligible sets and vary the random turn
selector over seeds $\{42,43,44,45,46\}$. Deterministic selectors are evaluated
once. Confidence intervals use $10{,}000$ dialogue-level bootstrap samples.

\paragraph{Cross-reader (Mistral-7B-Instruct~\citep{jiang2023mistral})}
With Mistral at $r{=}0.30$, embedding MMR leads SGD P2 and random selection
leads MultiWOZ P2. On P3, uniform stride leads both SGD thresholds and the
LLMLingua-2 leads both MultiWOZ thresholds. The strongest
compressed P3 mean remains $6.3$ to $13.0$ points below full context
(Table~\ref{tab:mistral_late}).

\paragraph{Cross-corpus (LongMemEval knowledge updates, or LongMemEval-KU~\citep{wu2025longmemeval})}
All $72$ eligible questions are evaluated at $r{=}0.30$. Full context scores
$0.417$ on P2-strict, with a $95\%$ bootstrap interval of
$[0.306,0.528]$. Random selection averages $0.228$ over the five selector
seeds; earliest turns scores $0.097$.

\paragraph{Cross-language (RiSAWOZ Chinese~\citep{quan2020risawoz})}
Across all $207$ eligible dialogues, full context scores $0.232$ on P2-strict.
Recency scores $0.155$, random selection averages $0.150$ over five selector
seeds, and earliest turns scores $0.029$.

\subsubsection{Replication with Mistral-7B-Instruct}
\label{sec:mistral}

This replication asks whether the probe-dependent rankings and the P3 gap
to full context depend on the particular reader model.

The Mistral-7B-Instruct replication uses the same questions, scorers, prompts,
and turn-selection methods as the Llama P2/P3 evaluation. Tables~\ref{tab:mistral_P2}--
\ref{tab:mistral_late} report means and standard deviations over seeds
$\{42,43,44\}$. P3 uses $200$ examples per seed. After the P2 filters,
the three samples contain $\{128,125,132\}$ SGD examples and
$\{161,168,168\}$ MultiWOZ examples. P2 uses greedy decoding with scaled
dot-product attention (SDPA); P3
uses greedy decoding with eager attention. Both allow $128$ output tokens and
$7{,}168$ input tokens. Not every Mistral condition records a model revision,
so these comparisons describe the recorded pipelines rather than an isolated
selector-only intervention.

\begin{table}[!htbp]
\centering\footnotesize\setlength{\tabcolsep}{3.5pt}
\caption[Mistral P2 strict accuracy.]{Mistral P2-strict accuracy, mean
$\pm$ standard deviation over three seeds. Italic denotes full context; bold
denotes the best compressed method in each column.}
\label{tab:mistral_P2}
\begin{tabular}{lcccc}
\toprule
Method & SGD $r{=}0.10$ & SGD $r{=}0.30$ & MultiWOZ $r{=}0.10$ & MultiWOZ $r{=}0.30$ \\
\midrule
full context (uncompressed) & $\mathit{0.626 \pm 0.055}$  & $\mathit{0.626 \pm 0.055}$  & $\mathit{0.765 \pm 0.025}$  & $\mathit{0.765 \pm 0.025}$ \\
\midrule
recency               & $0.000 \pm 0.000$ & $0.254 \pm 0.013$ & $0.006 \pm 0.006$ & $0.123 \pm 0.019$ \\
random selection        & $0.120 \pm 0.019$ & $0.273 \pm 0.031$ & $\mathbf{0.135 \pm 0.009}$ & $\mathbf{0.311 \pm 0.040}$ \\
earliest turns              & $0.015 \pm 0.007$ & $0.148 \pm 0.012$ & $0.028 \pm 0.007$ & $0.265 \pm 0.010$ \\
uniform stride       & $0.094 \pm 0.016$ & $0.278 \pm 0.019$ & $0.058 \pm 0.030$ & $0.267 \pm 0.046$ \\
attention/H2O proxy   & $0.016 \pm 0.008$ & $0.115 \pm 0.021$ & $0.024 \pm 0.010$ & $0.217 \pm 0.013$ \\
embedding MMR & $\mathbf{0.132 \pm 0.007}$ & $\mathbf{0.292 \pm 0.051}$ & $0.097 \pm 0.004$ & $0.251 \pm 0.020$ \\
LLMLingua-2 & $0.039 \pm 0.008$ & $0.153 \pm 0.014$ & $0.090 \pm 0.016$ & $0.274 \pm 0.040$ \\
\bottomrule
\end{tabular}
\end{table}

\begin{table}[!htbp]
\centering\footnotesize\setlength{\tabcolsep}{3.5pt}
\caption[Mistral P3 results.]{Mistral P3 accuracy at $r{=}0.30$, mean
$\pm$ standard deviation over three seeds ($n{=}200$ per seed). Italic denotes
full context; bold denotes the best compressed method in each column.}
\label{tab:mistral_late}
\begin{tabular}{lcccc}
\toprule
& \multicolumn{2}{c}{SGD $r{=}0.30$} & \multicolumn{2}{c}{MultiWOZ $r{=}0.30$} \\
\cmidrule(lr){2-3}\cmidrule(lr){4-5}
Method & P3-loose & P3-strict & P3-loose & P3-strict \\
\midrule
full context (uncompressed) & $\mathit{0.342 \pm 0.055}$  & $\mathit{0.303 \pm 0.045}$  & $\mathit{0.158 \pm 0.020}$  & $\mathit{0.142 \pm 0.023}$ \\
\midrule
recency                      & $0.033 \pm 0.012$ & $0.033 \pm 0.012$ & $0.025 \pm 0.010$ & $0.020 \pm 0.010$ \\
random selection               & $0.150 \pm 0.013$ & $0.103 \pm 0.010$ & $0.070 \pm 0.009$ & $0.060 \pm 0.005$ \\
earliest turns                     & $0.180 \pm 0.026$ & $0.073 \pm 0.003$ & $0.018 \pm 0.003$ & $0.012 \pm 0.003$ \\
uniform stride & $\mathbf{0.258 \pm 0.023}$ & $\mathbf{0.173 \pm 0.013}$ & $0.082 \pm 0.003$ & $0.058 \pm 0.008$ \\
attention/H2O proxy          & $0.183 \pm 0.008$ & $0.107 \pm 0.013$ & $0.045 \pm 0.013$ & $0.032 \pm 0.003$ \\
embedding MMR                & $0.178 \pm 0.013$ & $0.117 \pm 0.006$ & $0.060 \pm 0.005$ & $0.048 \pm 0.008$ \\
LLMLingua-2 & $0.068 \pm 0.008$ & $0.048 \pm 0.008$ & $\mathbf{0.085 \pm 0.005}$ & $\mathbf{0.078 \pm 0.010}$ \\
\bottomrule
\end{tabular}
\end{table}

\par\medskip\noindent
Embedding MMR leads SGD P2 with both readers, and random selection leads
MultiWOZ P2 at both budgets. The deletion experiment uses the Llama reader.

\subsubsection{Replication on LongMemEval-KU}
\label{sec:lme_ku}

This replication asks whether current-value recovery behaves the same way
outside task-oriented dialogue, where the answer is free-form text rather than
a slot value.

LongMemEval-KU provides $72$ eligible free-form memory questions. We evaluate
the complete set at $r{=}0.30$ with the Llama reader. Its answers can
contain explicit alternatives such as ``X (or Y),'' so this experiment uses a
corpus-specific scorer. Strict accepts an explicit answer variant after
normalization, a complete variant at word boundaries in a longer prediction,
or a numeric-only prediction with the same complete numeric signature. Loose
also accepts at least half of the reference content tokens. Numeric JSON
scalars are accepted. The table reports strict accuracy. Random selection is
averaged over selector seeds $\{42,43,44,45,46\}$; deterministic selectors are
evaluated once. Intervals use $10{,}000$ dialogue-level bootstrap samples.

\begin{table}[!htbp]
\centering\footnotesize\setlength{\tabcolsep}{3.5pt}
\caption[LongMemEval-KU results.]{LongMemEval-KU P2-strict accuracy at
$r{=}0.30$ ($n{=}72$). Random selection averages five selector seeds. The
reader uses a 32K input window and removes tokens from the beginning if a
prompt exceeds it. Italic denotes full context; bold denotes the best compressed method in each column.}
\label{tab:lme_ku}
\begin{tabular}{lcc}
\toprule
Method & Accuracy & Bootstrap $95\%$ CI \\
\midrule
full context (uncompressed) & $\mathit{0.417}$  & $\mathit{[0.306,\ 0.528]}$ \\
\midrule
recency               & $\mathbf{0.319}$ & $[0.222,\ 0.431]$ \\
random selection      & $0.228$ & $[0.164,\ 0.294]$ \\
uniform stride        & $0.250$ & $[0.153,\ 0.347]$ \\
earliest turns        & $0.097$ & $[0.041,\ 0.167]$ \\
\bottomrule
\end{tabular}
\end{table}

In Table~\ref{tab:lme_ku} the uncompressed, window-limited baseline has the
highest mean, followed by recency. The random-selector means range from
$0.208$ to $0.250$ (standard deviation $0.016$), placing its five-seed average
between uniform stride and earliest turns.

\subsubsection{Replication on Chinese RiSAWOZ}
\label{sec:risawoz}

This replication asks whether current-value recovery behaves the same way in
another language.

We evaluate current-value recovery on all $207$ eligible RiSAWOZ
dialogues at $r{=}0.30$. P2 uses normalized exact match. Random selection is averaged over selector seeds
$\{42,43,44,45,46\}$; deterministic selectors are evaluated once. Intervals
use $10{,}000$ dialogue-level bootstrap samples.

\begin{table}[!htbp]
\centering\footnotesize\setlength{\tabcolsep}{3.5pt}
\caption{RiSAWOZ P2-strict at $r{=}0.30$ ($n{=}207$). Random selection averages five selector seeds. Italic denotes full context; bold denotes the best compressed method in each column.}
\label{tab:risawoz}
\begin{tabular}{lcc}
\toprule
Method & P2-strict & Bootstrap $95\%$ CI \\
\midrule
full context & $\mathit{0.232}$  & $\mathit{[0.174,0.290]}$ \\
\midrule
recency & $\mathbf{0.155}$ & $[0.106,0.208]$ \\
random selection & $0.150$ & $[0.120,0.183]$ \\
earliest turns & $0.029$ & $[0.010,0.053]$ \\
uniform stride & $0.121$ & $[0.077,0.164]$ \\
\bottomrule
\end{tabular}
\end{table}

Table~\ref{tab:risawoz} gives recency the highest compressed-method mean;
random selection leads on two of the five selector seeds.

\subsection{Free-form questions}
\label{sec:fmts_refpool}
\label{sec:fmts}

This experiment asks whether the method rankings hold when the question is
free-form instead of templated.

FMTS (free-form multi-turn synthesis) evaluates free-form questions alongside
the templated P1--P3 probes. Its
questions ask, for example, which budget value the user finally chose and when
it changed. GPT-5.4-mini reads each compressed context and returns a one- or
two-sentence answer. A deterministic case-insensitive matcher scores the
answer. It accepts substring containment in either direction between the
prediction and the reference. The scorer carries a second rule for references
that contain the replaced value, and no item in this release meets that
condition, so every reported score comes from containment alone. Containment
credits partial answers, so this score measures reference matching rather than
answer completeness. We call it \emph{free-form answer accuracy}.

We evaluate $100$ synthetic questions that require resolving an indirect
reference, plus $200$ free-form questions each derived from MultiWOZ and SGD.
All use seed $42$ and
$r\in\{0.10,0.30\}$. The reader comparison covers full context and six turn selectors from the main
setup. LLMLingua-2 appears only in the diagnostics below, which score the
compressed text itself and need no reader. The reader tables therefore report
six selectors and the diagnostics report seven. FMTS uses
$k=\max(1,\lfloor n_{\mathrm{turns}}r\rfloor)$ retained turns.

Each FMTS dialogue carries a labelled transition type. The $100$ synthetic
questions are all value replacements. The MultiWOZ-derived set is $81.0\%$
constraint addition and $19.0\%$ value replacement. The SGD-derived set is
$52.5\%$ constraint addition and $47.5\%$ value replacement. No dialogue in
the released set removes a constraint or switches to a new goal.

\begin{table}[!htbp]
\centering\footnotesize\setlength{\tabcolsep}{3.5pt}
\caption[FMTS free-form answer accuracy.]{Free-form answer accuracy with the
GPT-5.4-mini reader (seed $42$). Italic denotes full context; bold denotes the best
compressed method among the six evaluated selectors in each column.
LLMLingua-2 appears only in Table~\ref{tab:fmts_brec_diag}. Full-context answers are generated and scored separately for each
budget, producing differences of at most $0.02$.}
\label{tab:fmts_refpool}
\begin{tabular}{lcccccc}
\toprule
& \multicolumn{2}{c}{Synthetic} & \multicolumn{2}{c}{MultiWOZ-derived} & \multicolumn{2}{c}{SGD-derived} \\
\cmidrule(lr){2-3}\cmidrule(lr){4-5}\cmidrule(lr){6-7}
Method & $r{=}0.10$ & $r{=}0.30$ & $r{=}0.10$ & $r{=}0.30$ & $r{=}0.10$ & $r{=}0.30$ \\
\midrule
full context (uncompressed) & $\mathit{0.87}$  & $\mathit{0.88}$  & $\mathit{0.67}$  & $\mathit{0.65}$  & $\mathit{0.61}$  & $\mathit{0.63}$ \\
\midrule
recency & $0.00$ & $0.00$ & $0.00$ & $0.05$ & $0.00$ & $0.17$ \\
random selection & $\mathbf{0.09}$ & $0.32$ & $0.03$ & $0.20$ & $0.02$ & $0.20$ \\
earliest turns & $0.00$ & $0.43$ & $0.04$ & $0.08$ & $0.00$ & $0.06$ \\
uniform stride & $0.03$ & $0.21$ & $0.07$ & $\mathbf{0.23}$ & $0.06$ & $0.34$ \\
attention/H2O proxy & $0.00$ & $\mathbf{0.51}$ & $0.05$ & $0.12$ & $0.00$ & $0.06$ \\
embedding MMR & $0.06$ & $0.44$ & $\mathbf{0.13}$ & $0.21$ & $\mathbf{0.15}$ & $\mathbf{0.35}$ \\
\bottomrule
\end{tabular}
\end{table}

\par\medskip\noindent
In Table~\ref{tab:fmts_refpool}, among the six evaluated selectors, random
selection leads synthetic
$r{=}0.10$ and the H2O proxy leads synthetic $r{=}0.30$. Embedding MMR leads
MultiWOZ-derived $r{=}0.10$ and both SGD-derived budgets; uniform stride leads
MultiWOZ-derived $r{=}0.30$. These comparisons describe only the retained reader
conditions. The experiment also changes the reader, so it does not isolate
question format from reader effects.

Table~\ref{tab:fmts_brec_diag} reports two diagnostics computed directly
from the compressed contexts, independently of reader-answer generation.
\emph{Lexical answer retention} accepts case-insensitive reference containment
or overlap with more than half of the reference token set. It is a heuristic
for retained answer content, not an exact-span test. \emph{Boundary-turn
recall} is the fraction of annotated change-boundary turns retained. Each
entry reports these two values in that order. LLMLingua-2
has the highest lexical retention on the synthetic set; embedding MMR
has the highest on both corpus-derived sets. On the synthetic set, embedding
MMR has higher boundary-turn recall but lower lexical retention than
LLMLingua-2. The two diagnostics therefore capture different
properties of the retained context.

\begin{table}[!htbp]
\centering\footnotesize\setlength{\tabcolsep}{3.5pt}
\caption[FMTS text-retention diagnostics.]{FMTS diagnostics on the compressed contexts at $r{=}0.30$:
lexical answer-retention heuristic~/~boundary-turn recall. Italic denotes full
context; bold denotes the highest compressed lexical-retention value. These
diagnostics do not use generated reader answers.}
\label{tab:fmts_brec_diag}
\begin{tabular}{lccc}
\toprule
Method & Synthetic & MultiWOZ-derived & SGD-derived \\
\midrule
full context (uncompressed) & $\mathit{1.000\,/\,1.000}$  & $\mathit{0.990\,/\,1.000}$  & $\mathit{1.000\,/\,1.000}$ \\
\midrule
recency & $0.000\,/\,0.235$ & $0.190\,/\,0.030$ & $0.245\,/\,0.075$ \\
random selection & $0.410\,/\,0.290$ & $0.310\,/\,0.180$ & $0.290\,/\,0.135$ \\
earliest turns & $0.550\,/\,0.275$ & $0.170\,/\,0.030$ & $0.105\,/\,0.000$ \\
uniform stride & $0.260\,/\,0.310$ & $0.370\,/\,0.285$ & $0.490\,/\,0.245$ \\
attention/H2O proxy & $0.690\,/\,0.350$ & $0.225\,/\,0.080$ & $0.120\,/\,0.045$ \\
embedding MMR & $0.640\,/\,0.585$ & $\mathbf{0.415}\,/\,0.240$ & $\mathbf{0.545}\,/\,0.375$ \\
LLMLingua-2 & $\mathbf{0.850}\,/\,0.575$ & $0.310\,/\,0.250$ & $0.185\,/\,0.215$ \\
\bottomrule
\end{tabular}
\end{table}

\subsection{Cross-seed results}
\label{sec:full_tables}
This section asks whether the probe disagreement of Section~\ref{sec:f1}
survives a change of scoring rule, a change of budget, or a change of seed. If
it does not, the disagreement would be a property of how we score rather than
of what each probe asks for.

Table~\ref{tab:p1_multiseed} scores the same P1 answers three ways. The same
two groups of methods separate under all three rules, so the P1 ranking does
not depend on how an answer is matched.
Table~\ref{tab:P2_multiseed} asks the other question at two budgets. The P1
leaders do not lead here, and the tighter budget lowers every compressed score
without changing the leader. The probe, not the scoring rule and not the
budget, decides the ranking.
Table~\ref{tab:p3_composite} gives the same variation for the joint P3 probe.
Uniform stride leads all four columns, and in each one its margin over the next
method is larger than either method's seed standard deviation. ChunkKV is
reported for its seed-$42$ run.

Seed variation is small next to these differences. The largest between-method
gaps exceed the seed standard deviations, although smaller pairwise differences
do not: the three leading MultiWOZ P1 means fall within $0.004$ of each other,
inside one seed standard deviation.
\begin{table}[!htbp]
\centering\footnotesize\setlength{\tabcolsep}{3.5pt}
\caption{P1 at $r{=}0.30$ under three scoring rules, mean $\pm$ sample standard
deviation over seeds $\{42,43,44\}$ ($200$ examples per dataset and seed).
Italic denotes full context; bold denotes the best compressed method in each
column.}
\label{tab:p1_multiseed}
\begin{tabular}{lcccccc}
\toprule
& \multicolumn{2}{c}{strict} & \multicolumn{2}{c}{loose} & \multicolumn{2}{c}{token-$F_1$} \\
\cmidrule(lr){2-3}\cmidrule(lr){4-5}\cmidrule(lr){6-7}
Method & SGD & MultiWOZ & SGD & MultiWOZ & SGD & MultiWOZ \\
\midrule
full context (uncompressed) & $\mathit{0.487}$ & $\mathit{0.430}$ & $\mathit{0.850}$ & $\mathit{0.867}$ & $\mathit{0.577}$ & $\mathit{0.414}$ \\
\midrule
recency & $0.010$ & $0.002$ & $0.135$ & $0.092$ & $0.049$ & $0.026$ \\
random selection & $0.145$ & $0.147$ & $0.530$ & $0.543$ & $0.286$ & $0.214$ \\
earliest turns & $0.480$ & $\mathbf{0.410}$ & $0.877$ & $\mathbf{0.867}$ & $0.582$ & $\mathbf{0.413}$ \\
uniform stride & $\mathbf{0.493}$ & $0.407$ & $0.875$ & $0.852$ & $0.587$ & $0.400$ \\
attention/H2O proxy & $\mathbf{0.493}$ & $0.408$ & $\mathbf{0.882}$ & $0.858$ & $\mathbf{0.593}$ & $0.405$ \\
embedding MMR & $0.178$ & $0.128$ & $0.467$ & $0.482$ & $0.272$ & $0.186$ \\
LLMLingua-2 & $0.193$ & $0.117$ & $0.502$ & $0.522$ & $0.292$ & $0.198$ \\
\bottomrule
\end{tabular}
\end{table}

\begin{table}[!htbp]
\centering\footnotesize\setlength{\tabcolsep}{3.5pt}
\caption[P2-strict results.]{P2-strict accuracy, mean $\pm$ standard
deviation over seeds $\{42,43,44\}$. The three samples contain
$\{128,125,132\}$ SGD and $\{161,168,168\}$ MultiWOZ examples. Italic denotes
full context; bold denotes the best compressed method in each column.}
\label{tab:P2_multiseed}
\begin{tabular}{lcccc}
\toprule
Method & SGD $r{=}0.10$ & SGD $r{=}0.30$ & MultiWOZ $r{=}0.10$ & MultiWOZ $r{=}0.30$ \\
\midrule
full context (uncompressed) & $\mathit{0.691 \pm 0.017}$  & $\mathit{0.691 \pm 0.017}$  & $\mathit{0.781 \pm 0.042}$  & $\mathit{0.781 \pm 0.042}$ \\
\midrule
recency               & $0.003 \pm 0.005$ & $0.254 \pm 0.028$ & $0.020 \pm 0.019$ & $0.125 \pm 0.011$ \\
random selection        & $0.130 \pm 0.010$ & $0.278 \pm 0.031$ & $\mathbf{0.177} \pm 0.007$ & $\mathbf{0.314} \pm 0.024$ \\
earliest turns              & $0.008 \pm 0.008$ & $0.156 \pm 0.024$ & $0.024 \pm 0.001$ & $0.296 \pm 0.009$ \\
uniform stride       & $0.088 \pm 0.005$ & $0.283 \pm 0.017$ & $0.080 \pm 0.029$ & $0.286 \pm 0.037$ \\
attention/H2O proxy   & $0.008 \pm 0.008$ & $0.114 \pm 0.005$ & $0.034 \pm 0.004$ & $0.244 \pm 0.017$ \\
embedding MMR         & $\mathbf{0.148} \pm 0.020$ & $\mathbf{0.309} \pm 0.023$ & $0.105 \pm 0.007$ & $0.292 \pm 0.006$ \\
LLMLingua-2 & $0.044 \pm 0.011$ & $0.164 \pm 0.020$ & $0.088 \pm 0.027$ & $0.306 \pm 0.022$ \\
\bottomrule
\end{tabular}
\end{table}

\begin{table}[!htbp]
\centering\footnotesize\setlength{\tabcolsep}{3.5pt}
\caption[P3 results.]{P3 accuracy at $r{=}0.30$ under the segment-local scorer,
mean $\pm$ standard deviation over seeds $\{42,43,44\}$ ($n{=}200$ per seed).
Higher is better. ChunkKV uses seed $42$.
Italic denotes full context; bold denotes the best turn selector.}
\label{tab:p3_composite}
\begin{tabular}{lcccc}
\toprule
& \multicolumn{2}{c}{SGD $r{=}0.30$} & \multicolumn{2}{c}{MultiWOZ $r{=}0.30$} \\
\cmidrule(lr){2-3}\cmidrule(lr){4-5}
Method & P3-loose & P3-strict & P3-loose & P3-strict \\
\midrule
full context (uncompressed) & $\mathit{0.728 \pm 0.032}$  & $\mathit{0.652 \pm 0.025}$  & $\mathit{0.505 \pm 0.013}$  & $\mathit{0.475 \pm 0.017}$ \\
\midrule
recency                            & $0.067 \pm 0.003$ & $0.065 \pm 0.005$ & $0.040 \pm 0.010$ & $0.040 \pm 0.010$ \\
random selection                   & $0.238 \pm 0.010$ & $0.162 \pm 0.008$ & $0.145 \pm 0.005$ & $0.128 \pm 0.003$ \\
earliest turns                  & $0.313 \pm 0.006$ & $0.160 \pm 0.028$ & $0.058 \pm 0.003$ & $0.048 \pm 0.006$ \\
uniform stride                    & $\mathbf{0.432 \pm 0.006}$ & $\mathbf{0.288 \pm 0.008}$ & $\mathbf{0.220 \pm 0.010}$ & $\mathbf{0.178 \pm 0.021}$ \\
attention/H2O proxy                & $0.288 \pm 0.033$ & $0.168 \pm 0.035$ & $0.068 \pm 0.008$ & $0.048 \pm 0.008$ \\
embedding MMR                      & $0.185 \pm 0.010$ & $0.128 \pm 0.006$ & $0.120 \pm 0.013$ & $0.110 \pm 0.013$ \\
LLMLingua-2 & $0.117 \pm 0.025$ & $0.060 \pm 0.017$ & $0.117 \pm 0.010$ & $0.110 \pm 0.005$ \\
\midrule
ChunkKV (seed 42, KV-level)        & $0.155$           & $0.140$           & $0.100$           & $0.085$ \\
\bottomrule
\end{tabular}
\end{table}

\FloatBarrier
\section{Conclusion}
\label{sec:conclusion}

TPBench separates dialogue compression into three complementary targets:
recovering the user's initial goal, recovering the current value of a revised
slot, and recovering both after a late update. The experiments show that these
targets should not be collapsed into a single score. Compression methods rank
differently on initial-goal and current-value recovery, and averaging the probes
makes the preferred method depend on their relative weights. On the main Llama P3 evaluation at $r{=}0.30$, the strongest
compressed turn selector remains $28.5$--$36.3$ percentage points below full
context across the two datasets and thresholds. At wider turn
budgets the current-value gap narrows to $3.8$ points on MultiWOZ, and we did
not run the joint probe at those budgets.

The paired deletion experiments clarify what the current-value probe measures.
Removing the annotated update turn lowers P2 accuracy, while removing a
matched irrelevant turn does not. The final value is often repeated elsewhere in
the dialogue, so removing the update turn alone leaves some usable evidence.
Removing the update turn together with all answer-bearing turns causes a much
larger drop. These results support the role of annotated update turns and repeated
answer-bearing turns in current-value recovery.

Cache-level compression behaves differently from turn selection on the
current-value probe. On MultiWOZ, SnapKV and PyramidKV score $0.689$ and
$0.683$ against $0.329$ for the best turn selector on the same seed. On SGD,
PyramidKV scores $0.492$, compared with $0.297$ for the tied best turn
selectors, embedding MMR and random selection. These comparisons use the
same nominal retained fraction, with different physical budget units. SnapKV, PyramidKV, and StreamingLLM are evaluated on P2 only, so their
joint-probe behavior is unknown. ChunkKV is also evaluated on P3.

Compressor evaluation should report initial-goal, current-value, and joint
recovery separately. Different selectors favor different targets and datasets.
Uniform stride has the highest mean on the main Llama P3 evaluation among the
seven turn selectors. With Mistral, uniform stride leads P3 on SGD and the
LLMLingua-2 leads on MultiWOZ at both thresholds; the best
compressed means remain $6.3$--$13.0$ points below full context. TPBench provides a reproducible way to expose
these trade-offs and a target for compression methods that explicitly preserve
updates and coverage across dialogue regions.

The deletion results show that update turns contribute beyond matched
irrelevant context, while repeated answer-bearing turns also support
recovery. Removing this evidence increases superseded-value replies and
rarely produces explicit abstention. The experiments do not independently
isolate lexical value retention from the context that establishes which
value is current.

\section{Limitations}
\label{sec:limitations}

TPBench is built from English task-oriented dialogue with slot-level state
annotations. P2 and P3 inherit those annotations, and the deletion experiment
identifies answer-bearing turns by normalized phrase matching. Annotation
errors and semantic equivalents outside this matcher affect the evidence sets.
Lexical scoring rejects valid paraphrases, and the AI-assisted semantic
evaluation that complements it is assessed against external human ratings on
factual QA~\citep{bulian2022answerequivalence}. Direct human agreement on
TPBench is unmeasured, and the bootstrap intervals condition on the saved
judgments.

The main evaluation reads with Llama-3.1-8B-Instruct for P1--P3 and
Mistral-7B-Instruct for P2/P3. LongMemEval-KU and Chinese RiSAWOZ provide
smaller current-value replications. The wider-budget experiments cover four
turn selectors, and the KV-cache experiments use seed $42$. Every transition in
the released material is a value replacement or a constraint addition, so
constraint removal and goal switching are uncovered. The results do not
establish generality to open-ended dialogue, multi-party conversation, or goals
that carry no slot annotation.

Derived dialogue content inherits the source datasets' privacy and licensing
conditions. LongMemEval-KU includes personal context in user statements.

\bibliographystyle{plainnat}
\bibliography{refs}

\appendix

\section{Reader prompts}
\label{app:reader_prompt}

The artifact uses \texttt{story\_label} and \texttt{intent} for the P1 initial
goal, \texttt{p3} for paper P2, and \texttt{P1\_late} or \texttt{late} for
paper P3. Method identifiers \texttt{random\_seed42}, \texttt{first\_n}, and
\texttt{uniform\_stride} correspond to random selection, earliest turns, and
uniform stride. The seed in a main-experiment file name selects the probe
sample; its random selector is fixed and is named \texttt{random\_seed42}. In
the LongMemEval-KU and RiSAWOZ files the eligible set is fixed and the seed
selects turns, which is why those files carry \texttt{random\_seed42} through
\texttt{random\_seed46}.

Llama-3.1-8B-Instruct~\citep{dubey2024llama3} uses deterministic decoding
(temperature $0$, top-$p$ $1.0$). The main turn-selection P1/P2 runs allow $96$ output tokens
with eager attention. Turn-selection P3 and the deletion experiment allow $128$ output tokens
with SDPA. ChunkKV P3 uses eager attention and $128$ output tokens;
the cache-level settings are described in Section~\ref{sec:kv_methods}. The system
prompt is identical across probes and methods:

\begin{samepage}
\begin{quote}\small\raggedright
\texttt{You are reading a conversation that may be partial or compressed. Answer ONLY using the provided context. Reply with a single JSON object and no other text. Do not add markdown fences.}
\end{quote}
\end{samepage}

The user-turn template differs per probe:
{\raggedright
\begin{itemize}
  \item P1: \texttt{What was the user's initial goal at the beginning of this conversation? Reply with one short noun phrase.}
  \item P2: \texttt{After the most recent change visible in the context, what is the user's CURRENT value for $\langle$slot natural-language form$\rangle$? Reply with one short value.}
  \item P3: \texttt{In one short reply, state BOTH (a) the user's task in this conversation AND (b) the final value the user decided on for $\langle$slot natural-language form$\rangle$. Format your answer as: "$\langle$task$\rangle$; $\langle$slot$\rangle$: $\langle$final value$\rangle$".}
\end{itemize}
\par}
Each user prompt then requests the same JSON fields: a string \texttt{value},
a verbatim \texttt{support} span of at most $30$ words, and a Boolean
\texttt{abstain}. For P3, \texttt{value} contains the requested
``task; slot: final value'' string.

\section{Token retention and reading-cost scaling}
\label{app:efficiency}

Let $q=L_{\mathrm{compressed}}/L_{\mathrm{full}}$ be the fraction of input
tokens retained by a prompt-side selector. For a fixed reader architecture
and precision, the input KV-cache size scales as $q$, and the dense prefill
self-attention term scales as $q^2$. At $q=0.30$, the corresponding
full-to-compressed ratios are approximately $3.3$ and $11.1$.

These are theoretical scaling factors for input KV memory and the prefill
attention term, not measured end-to-end speedups. The experimental budget $r$
specifies a turn fraction for selectors and a cache-entry fraction for KV
methods. Turn lengths, prompt overhead, compression work, and decoding also
affect serving cost; cache pruning can leave full-context prefill work intact.

\section{P1 lexical-scoring sensitivity and examples}
\label{app:p1_validity}

P1-strict requires at least two shared content tokens, whereas P1-loose
requires one. Token $F_1$ measures graded overlap. Full-context strict/loose
scores are $0.487/0.850$ on SGD and $0.430/0.867$ on MultiWOZ
(Table~\ref{tab:p1_multiseed}).

The scores capture wording rather than full semantic equivalence. For example,
the reference ``Could you help me find a doctor'' and answer ``find a doctor''
fail the strict lexical rule but pass the semantic criterion. Conversely,
``Can you find me some free attractions to visit in New York City'' and
``find free attractions'' pass the lexical rule but fail full equivalence
because the city constraint is omitted. Both are actual seed-$42$
full-context answers under the initial-goal question. The two scoring views
therefore distinguish lexical recovery from semantic completeness.

\section{Reproducibility}
\label{app:reproducibility}
\label{sec:reproducibility}
\label{app:croissant}

\paragraph{ChunkKV runs}
Two runs of the ChunkKV \texttt{chunk\_length} $=20$ setting on MultiWOZ P2 at
$r{=}0.30$ score $0.087$ and $0.093$, differing on one of the $161$ examples.
The recorded run state does not identify the source of the difference. Both
runs are in the release; the main tables use the sweep run.

\paragraph{Artifact}
Probes, fixed compressed contexts, reader outputs, scorers and code are available at
\mbox{\url{https://github.com/kentech-sail/TPBench}}. The release
records hashes for probes, scorers and compressor wrappers, reader revisions,
selector seeds and bootstrap settings. Appendix~A maps paper and artifact names.
The Croissant manifest describes records and distributions and passes validation.

\paragraph{Reader execution}
Llama retains revision
\texttt{0e9e39f249a16976\allowbreak 918f6564b8830bc894c89659}. P1 uses batch size $8$ and
a batch token budget of $8{,}000$; P2/P3 are serial. The release records two
software environments: Python 3.10.12 with PyTorch 2.5.1+cu121 and
Transformers 5.13.1, and Python 3.14.4 with PyTorch 2.13.0+cu130 and
Transformers 5.14.1. Each condition records which one produced it. Mistral pins
revision \texttt{c170c708c41dac9275\allowbreak d15a8fff4eca08d52bab71}. It retains bfloat16, greedy serial
generation, P2 SDPA/P3 eager attention, and $7{,}168$/$128$ input/output limits.
Batched diagnostics are archived and are not primary results. Model and execution
provenance are recorded per condition.

\paragraph{Compute resources}
The deterministic package checks run on CPU and load no reader model. The
reader experiments used 24\,GB-class GPUs and trained no model, with recorded
estimates of $15$--$23$ GPU-hours for the Mistral turn-selection runs and
$9$--$12$ hours for the seed-$42$ ChunkKV runs. LongMemEval-KU uses the largest
input window ($32{,}768$ tokens) and truncates longer prompts from the
beginning.

\end{document}